\documentclass[10pt,twocolumn,letterpaper]{article}

\usepackage{cvpr}              

\usepackage{graphicx}
\usepackage{amsmath}
\usepackage{amssymb}
\usepackage{booktabs}
\usepackage[pagebackref,breaklinks,colorlinks]{hyperref}

\usepackage{float}
\usepackage{mathtools}
\usepackage{soul}
\usepackage{algorithm}
\usepackage{algorithmic}
\usepackage[export]{adjustbox}
\usepackage{pgfplots}
\usepackage{pgfplotstable}
\usepackage{datatool}
\usepackage{overpic}
\usepackage{listings}

\newif\ifdraft
\draftfalse

\ifdraft
\newcommand{\dcc}[1]{{\color{red}[\textbf{Danny:} #1]}}
\newcommand{\avc}[1]{{\color{purple}[\textbf{Andrey:} #1]}}
\newcommand{\ahc}[1]{{\color{orange}[\textbf{Amir:} #1]}}
\newcommand{\sfc}[1]{{\color{blue}[\textbf{Shlomi:} #1]}}

\newcommand{\drop}[1]{}

\else
\newcommand{\dcc}[1]{}
\newcommand{\avc}[1]{}
\newcommand{\sfc}[1]{}
\newcommand{\ahc}[1]{}

\fi

\newcommand{\ourmethod}{StyleAligned}

\newcommand{\imfeat}{\phi}

\newcommand{\queries}{Q}
\newcommand{\keys}{K}
\newcommand{\values}{V}
\newcommand{\queriesm}{$\queries$}
\newcommand{\keysm}{$\keys$}
\newcommand{\valuesm}{$\values$}
\newcommand{\shared}{1 \dots n}

\newcommand{\refim}{reference}
\newcommand{\tarim}{target}

\makeatletter
\DeclareRobustCommand\onedot{\futurelet\@let@token\@onedot}
\def\@onedot{\ifx\@let@token.\else.\null\fi\xspace}
\DeclareMathAlphabet\mathbfcal{OMS}{cmsy}{b}{n}

\makeatother

\makeatletter
\def\blfootnote{\xdef\@thefnmark{}\@footnotetext}
\makeatother

\newcommand{\reals}{\mathbb{R}}

\newif\ifwatermark
\watermarktrue
\draftfalse

\usepackage[pagebackref,breaklinks,colorlinks]{hyperref}
\newcommand*{\affaddr}[1]{#1} 
\newcommand*{\affmark}[1][*]{\textsuperscript{#1}}
\newcommand*\samethanks[1][\value{footnote}]{\footnotemark[#1]}

\usepackage[symbol]{footmisc}
\usepackage[accsupp]{axessibility} 

\usepackage[capitalize]{cleveref}

\def\cvprPaperID{11561} 
\def\confName{CVPR}
\def\confYear{2024}

\begin{document}

\title{Style Aligned Image Generation via Shared Attention}

\author{%
\large
Amir Hertz\footnote{}~ \affmark[1], Andrey Voynov\samethanks~ \affmark[1], Shlomi Fruchter\footnote{}~~\affmark[1], and Daniel Cohen-Or\samethanks{}~~\affmark[1,]\affmark[2]\\
\normalsize{\affaddr{\affmark[1] Google Research}}\\
\normalsize{\affaddr{\affmark[2] Tel Aviv University}}\\
}

\maketitle

\begin{abstract}
Large-scale Text-to-Image (T2I) models have rapidly gained prominence across creative fields, generating visually compelling outputs from textual prompts. However, controlling these models to ensure consistent style remains challenging, with existing methods necessitating fine-tuning and manual intervention to disentangle content and style. In this paper, we introduce \textit{\ourmethod{}}, a novel technique designed to establish style alignment among a series of generated images. By employing minimal `attention sharing' during the diffusion process, our method maintains style consistency across images within T2I models. This approach allows for the creation of style-consistent images using a reference style through a straightforward inversion operation. Our method's evaluation across diverse styles and text prompts demonstrates high-quality synthesis and fidelity, underscoring its efficacy in achieving consistent style across various inputs.

\end{abstract}
\footnotetext[1]{Equal contribution.}
\footnotetext[2]{Equal Advising.}

\renewcommand*{\thefootnote}{\arabic{footnote}}

\begin{figure}[t]
\footnotesize
\centering
    \includegraphics[width=1\columnwidth]{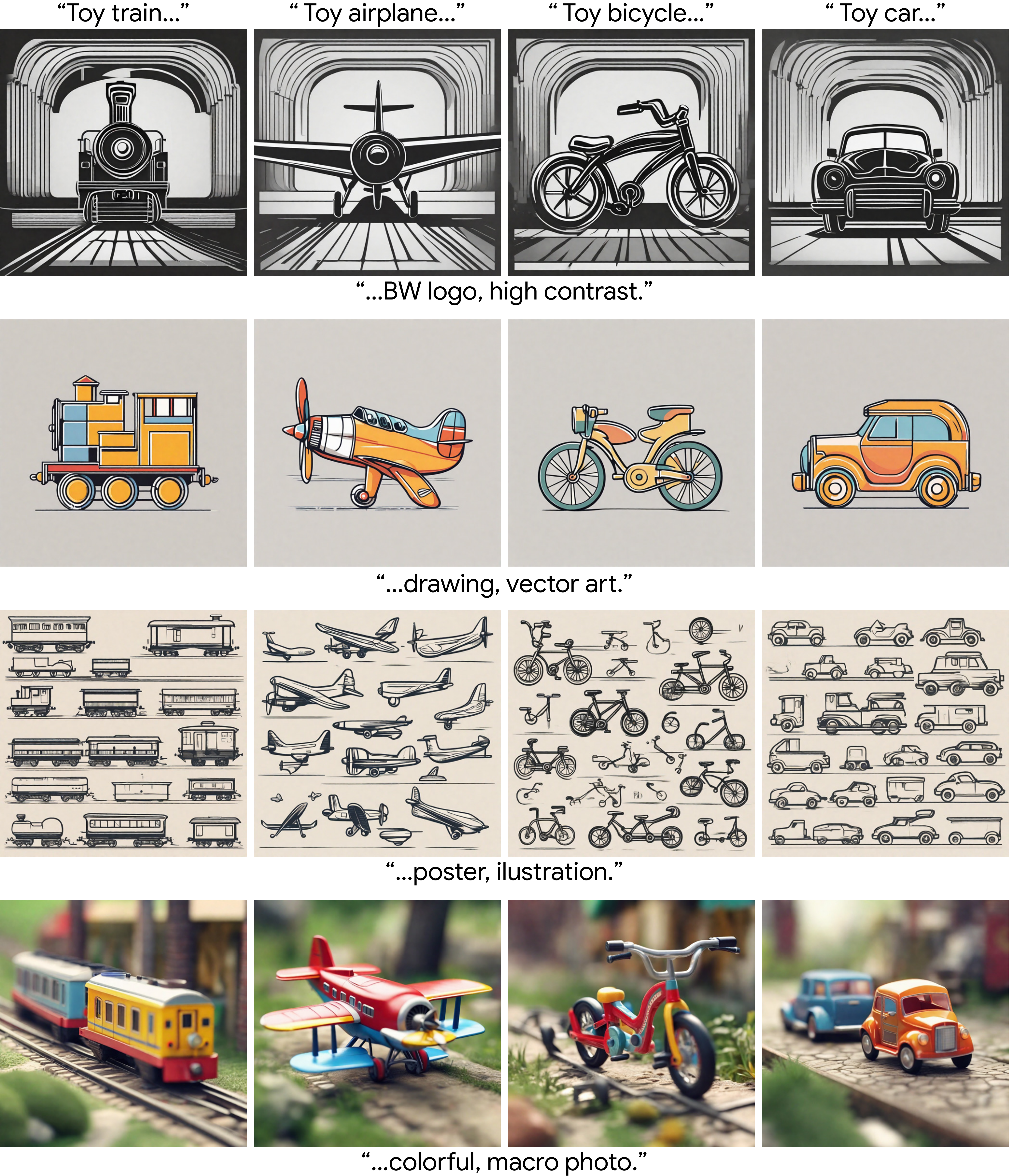}
\caption{\textbf{Style aligned image set generation.} \emph{By fusing the features of the toy train image (left) during the diffusion process, we can generate an image set of different content that shares the style.}}
\label{fig:teaser}
\end{figure}

\section{Introduction}

Large-scale Text-to-Image (T2I) generative models \cite{dalle, imagen, latent_diffusion} have emerged as an essential tool across creative disciplines, such as art, graphic design, animation, architecture, gaming and more.
These models show tremendous capabilities of translating an input text into an appealing visual result that is aligned with the input description. 

\begin{figure}[t]
\footnotesize
\centering
    \includegraphics[width=1\columnwidth]{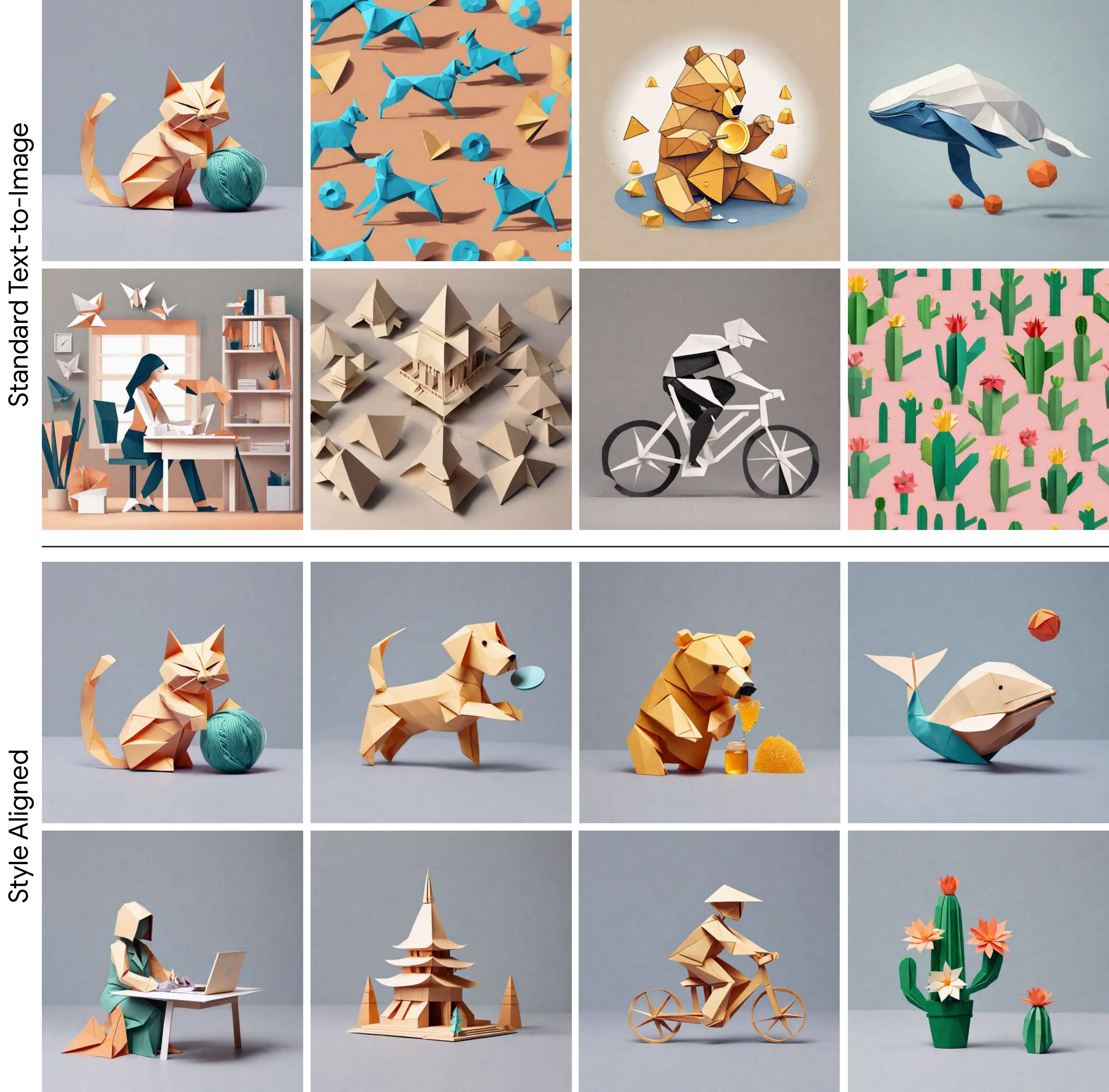}
\caption{\textbf{Standard text-to-image vs. StyleAligned set generation.} \emph{Given style description of ``minimal origami'', standard text-to-image generation (top) results with an unaligned image set while our method (bottom) can generate variety of style aligned content.}}
\label{fig:aligned}
\end{figure}

An envisioned application of T2I models revolves around the rendition of various concepts in a way that shares a consistent style and character, as though all were created by the same artist and method (see Fig. \ref{fig:teaser}). While proficient in aligning with the textual description of the style, state-of-the-art T2I models often create images that diverge significantly in their interpretations of the same stylistic descriptor, as depicted in Fig. \ref{fig:aligned}.

Recent methods mitigate this by fine-tuning the T2I model over a set of images that share the same style \cite{Gal2022AnII, sohn2023styledrop}. This optimization is computationally expensive and usually requires human input in order to find a plausible subset of images and texts that enables the disentanglement of content and style.

We introduce \ourmethod{}, a method that enables \textit{consistent style interpretation} across a set of generated images (Fig. \ref{fig:teaser}). Our method requires no optimization and can be applied to any attention-based text-to-image diffusion model. We show that adding minimal \textit{attention sharing} operations along the diffusion process, from each generated image to the first one in a batch, leads to a style-consistent set.
Moreover, using diffusion inversion, our method can be applied to generate style-consistent images given a reference style image, with no optimization or fine-tuning.

We present our results over diverse styles and text prompts, demonstrating high-quality synthesis and fidelity to the prompts and reference style. We show  diverse examples of generated images that share their style with a reference image that can  possibly be a given input image. 
Importantly, our technique stands as a zero-shot solution, distinct from other personalization techniques, as it operates without any form of optimization or fine-tuning.
For our code and more examples, please visit the project page \textit{\color{magenta}\url{style-aligned-gen.github.io}}

\section{Related Work}
\label{sec:related_work}

\textbf{Text-to-image generation.}
Text conditioned image generative models~\cite{nichol2021glide, Rombach2021HighResolutionIS, chang2023muse} show unprecedented capabilities of generating high quality images from text descriptions.
In particular, T2I diffusion models~\cite{Rombach2021HighResolutionIS, Podell2023SDXLIL, Saharia2022PhotorealisticTD} are pushing the state of the art and 
they are quickly adopted for different generative visual tasks like inpainting~\cite{avrahami2023blendedlatent, Saharia2021PaletteID}, image-to-image translation \cite{voynov2022sketch, Ye2023IPAdapterTC}, local image editing \cite{couairon2022diffedit, HubermanSpiegelglas2023}, subject-driven image generation~\cite{ruiz2023hyperdreambooth, Tewel2023KeyLockedRO} and more.

\textbf{Attention Control in diffusion models.}
Hertz et al.~\cite{hertz2022prompt} have shown how cross and self-attention maps within the diffusion process determine the layout and content of the generated images. Moreover, they showed how the attention maps can be used for controlled image generation. Other studies have leveraged modifications in attention layers to enhance the fidelity or diversity of generated images
~\cite{Chefer2023AttendandExciteAS, Patashnik2023LocalizingOS},  or apply attention control for image editing ~\cite{mokady2022null, pnpDiffusion2022, cao2023masactrl, parmar2023zero, epstein2023diffusion, park2024shape}. However, in contrast to prior approaches that primarily enable structure-preserving image editing, our method excels at generating images with diverse structures and content while maintaining a consistent style interpretation.

\textbf{Style Transfer.} Transferring a style from a reference image to a target content image is well studied subject in computer graphics. Classic works~\cite{efros2023image, hertzmann2023image, efros1999texture, lee2010directional} rely on optimization of handcrafted features and texture resampling algorithms from an input texture image, combined with structure constrains of a content image.
With the progress of deep learning research, another line of works utilizes deep neural priors for style transfer optimization using deep features of pre-trained networks~\cite{gatys2016image, tumanyan2022splicing}, or injecting attention features from a style image to a target one \cite{alaluf2023crossimage}.
More related to our approach, Huang et al.~\cite{huang2017arbitrary} introduced a real time style transfer network based on Adaptive Instance Normalization layers (AdaIN) that are used to normalize deep features of a target image using deep features statistics of a reference style image. Follow-up works, employ the AdaIN layer for additional unsupervised learning tasks, like style-based image generation~\cite{karras2019style} and Image2Image translation~\cite{huang2018multimodal, liu2019few}.

\textbf{T2I Personalization}
To generalize T2I over new visual concepts, several works developed different optimization techniques over a small collection of input images that share the same concept~\cite{Gal2022AnII, ruiz2023hyperdreambooth, Voynov2023PET, Han2023SVDiffCP}. In instances where the collection shares a consistent style, the acquired concept becomes the style itself, affecting subsequent generations. 
Most close to our work is StyleDrop~\cite{sohn2023styledrop}, a style personalization method that relies on fine-tuning of light weight adapter layers~\cite{houlsby2019parameter} at the end of each attention block in a non-autoregressive generative text-to-image transformer~\cite{chang2023muse}.
StyleDrop can generate a set of images in the same style of by training the adapter layers over a collection of images that share the same style. However, it struggles to generate a consistent image set of different content when training on a single image.

Our method can generate a consistent image set without optimization phase and without relying on several images for training. 
To skip the training phase, recent works developed dedicated personalization encoders \cite{gal2023encoder, Shi2023InstantBoothPT, Li2023BLIPDiffusionPS, Ye2023IPAdapterTC, Wei2023ELITEEV} that can directly inject new priors from a single input image to the T2I model. However, these methods encounter challenges to disentangle style from content as they focus on generating the same subject as in the input image.

\begin{figure*}[t]
\footnotesize
\centering
    \includegraphics[width=1.\textwidth]{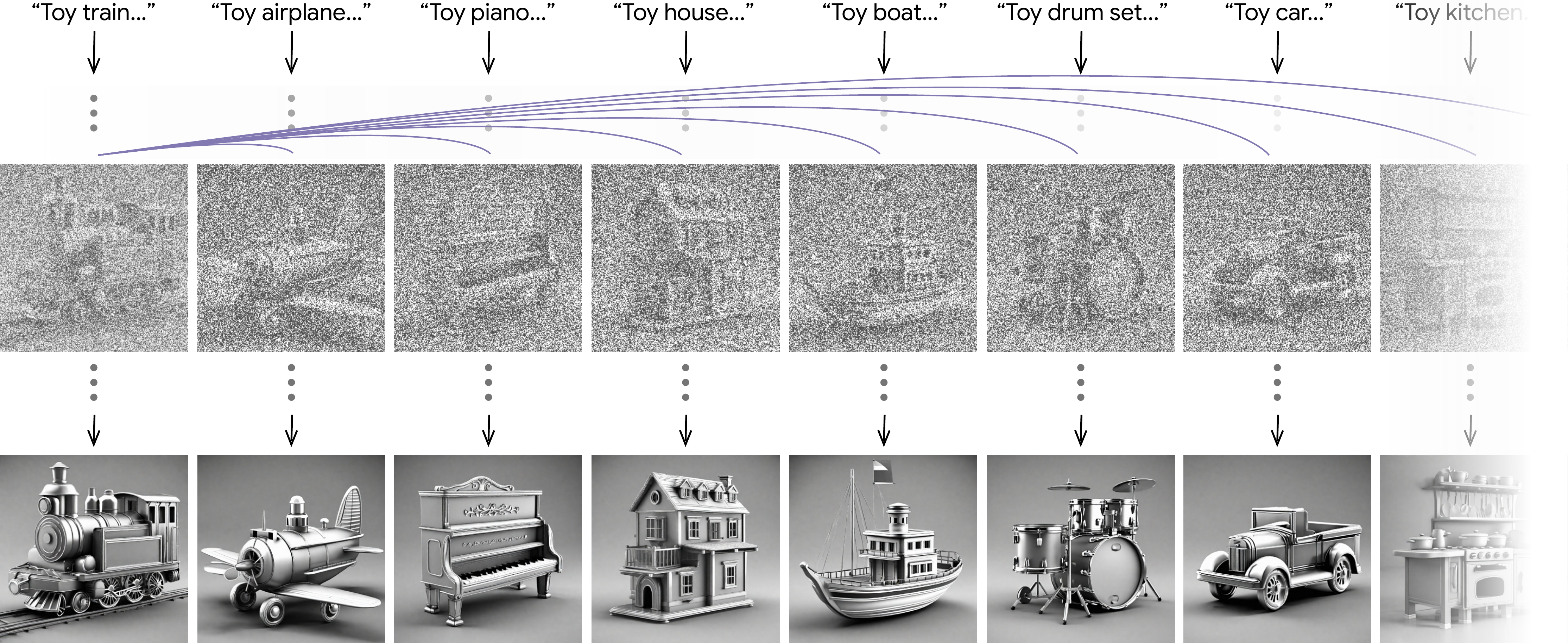}
\caption{\textbf{Style Aligned Diffusion.} \textit{Generation of images with a style aligned to the reference image on the left. In each diffusion denoising step all the images, except the reference, perform a shared self-attention with the reference image.}}

\label{fig:diagram}
\end{figure*}

\begin{figure}[t]
\footnotesize
\centering
    \includegraphics[width=.7\columnwidth]{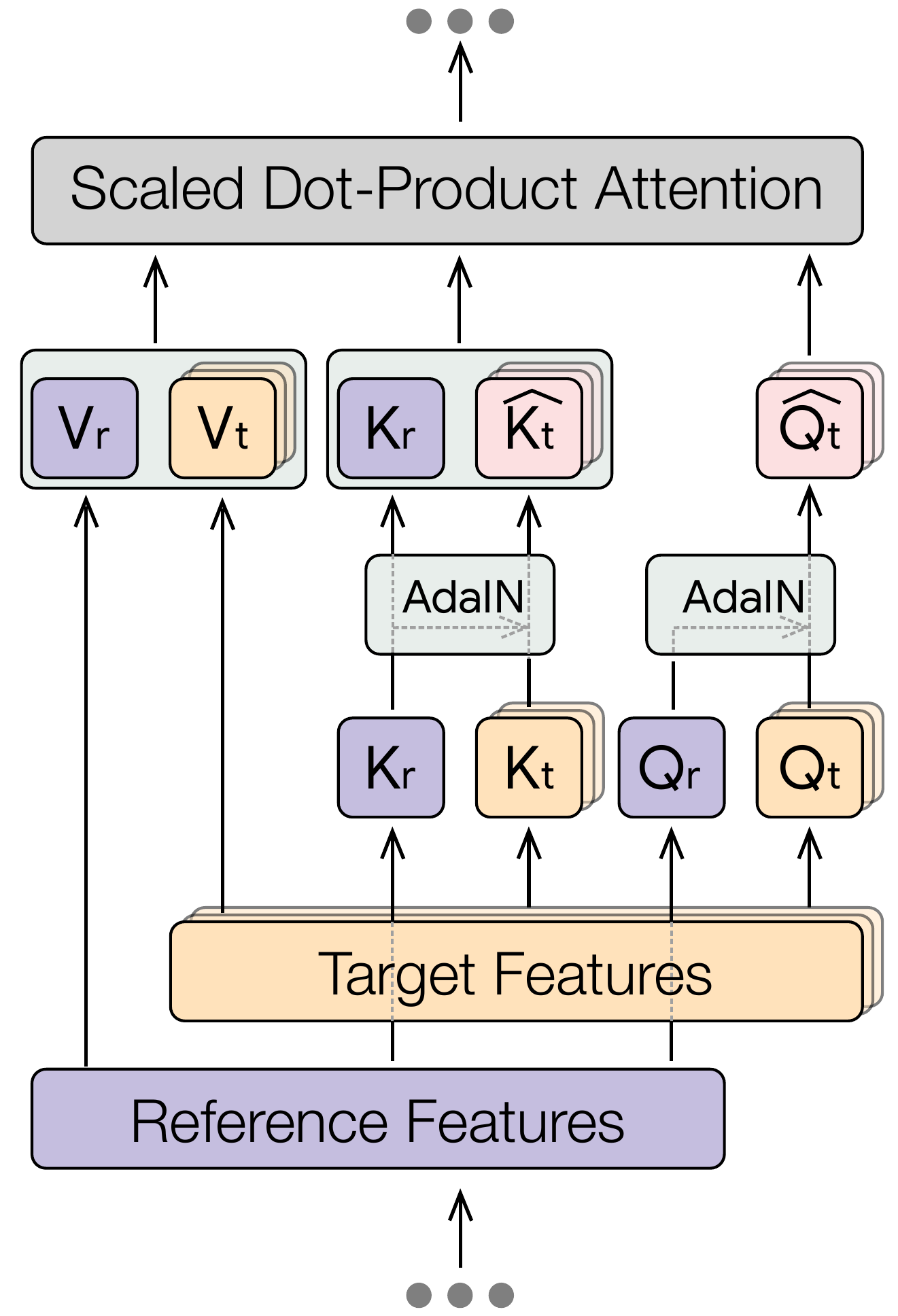}
\caption{\textbf{Shared attention layer.} \textit{The target images attends to the reference image by applying AdaIN over their queries and keys using the reference queries and keys respectively.
Then, we apply shared attention where the target features are updated by both  the target values $V_t$ and the reference  values $V_r$.}}
\label{fig:diagram-right-side}
\end{figure}

\begin{figure}[t]
\footnotesize
\centering
    \includegraphics[width=1\columnwidth]{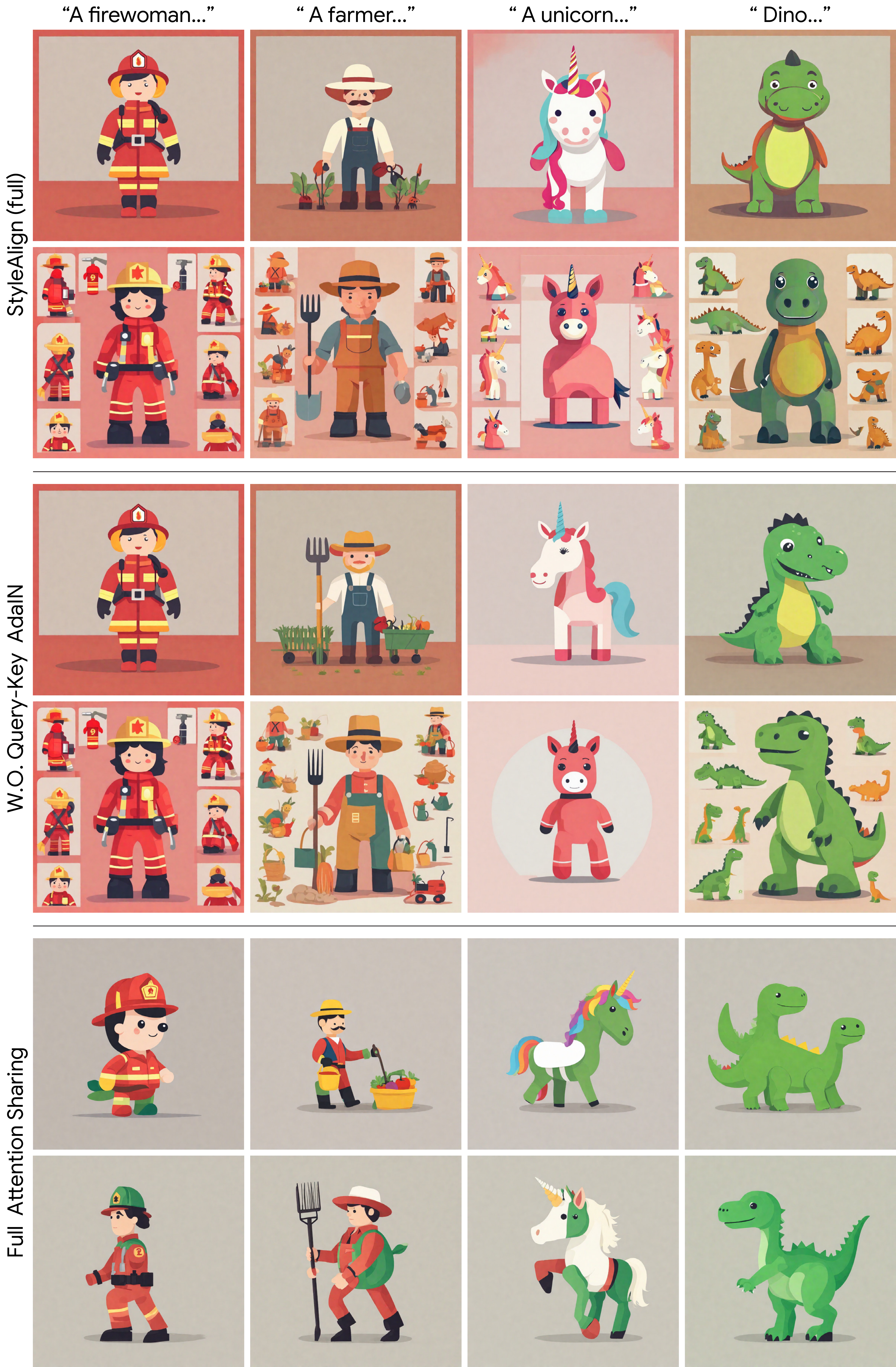}
\caption{\textbf{Ablation study -- qualitative comparison.} \emph{Each pair of rows shows two sets of images generated by the same set of prompts ``...in minimal flat design illustartion''  using different configurations of our method, and each row in a pair uses a different seed. Sharing the self--attention between all images in the set (bottom) results with some diversity loss (style collapse across many seeds) and content leakage within each set (colors from one image leak to another). Disabling the queries--keys AdaIN opeartion  results with less consistent image sets compared to our full method (top) which keeps on both diversity between different sets and consistency within each set.}}
\label{fig:ablation_qual}
\end{figure}

\section{Method overview}
In the following section we start with an overview of the  T2I diffusion process, and in particular the self--attention mechanism Sec.~\ref{preliminaries}.  We continue by presenting our attention-sharing operation within the self--attention layers that enable style aligned image set generation.

\subsection{Preliminaries}
\label{preliminaries}
Diffusion models~\cite{sohl2015deep, ho2020denoising} are generative latent variable models that aim to model a distribution $p_\theta(x_0)$ that approximates the data distribution $q(x_0)$ and are easy to sample from.
Diffusion models are trained to reverse the diffusion ``forward process'':
$$
x_t = \sqrt{\alpha_t}x_0+\sqrt{1-\alpha_t}\epsilon,~~\epsilon\sim N(0,I),
$$
where $t \in [0, \infty)$ and the values of $\alpha_t$ are determined by a scheduler such that $\alpha_0 = 1$ and  $\lim_{t\to\infty}\alpha_t = 0$.
During inference, we sample an image by gradually denoising an input noise image $x_T\sim \mathcal{N}(0,I)$ via the reverse process:
$$
x_{t-1} = \mu_{t-1} + \sigma_t z,~~z\sim N(0,I),
$$
where the value of $\sigma_t$ is determined by the sampler and $\mu_{t-1}$ is given by
$$
\mu_{t-1} =\dfrac{\sqrt{\alpha_{t-1}} x_t}{\sqrt{\alpha_t}} + \left( \sqrt{1 - \alpha_{t-1}}- \dfrac{\sqrt{1 - \alpha_t}}{\sqrt{\alpha_t}}\right) \epsilon_\theta(x_t,t), 
$$
where $\epsilon_\theta(x_t,t)$ is the output of a diffusion model parameterized by $\theta$.

Moreover, this process can be generalized for learning a marginal distribution using an additional input condition. That leads text-to-image diffusion models (T2I), where the output of the model $\epsilon_\theta(x_t,t, y)$ is conditioned on a text prompt $y$.

\textbf{Self-Attention in T2I Diffusion Models.} State-of-the-art T2I diffusion models \cite{Podell2023SDXLIL, Saharia2022PhotorealisticTD, BetkerImprovingIG} employ a U-Net architecture \cite{ronneberger2015u} that consists of convolution layers and transformer attention blocks~\cite{NIPS2017_3f5ee243}. In these attention mechanisms, deep image features $\imfeat \in \reals^{m \times d_h}$ attend to each other via self-attention layers and to contextual text embedding via cross-attention layers.

Our work operates at the self-attention layers where deep features are being updated by attending to each other. First, the features are projected into queries $\queries \in {m \times d_k}$, keys $\keys \in {m \times d_k}$ and values $\values \in {m \times d_h}$ through learned linear layers. Then, the attention is computed by the scaled dot-product attention:
$$
\textrm{Attention}(\queries, \keys, \values) = \textrm{softmax}\left(\dfrac{\queries\keys^T}{\sqrt{d_k}} \values\right),
$$
where $d_k$ is the dimension of \queriesm{} and \keysm. Intuitively, each image feature is updated by a weighted sum of \valuesm{}, where the weight depends on the correlation between the projected query $q$ and the keys \keysm.
In practice, each self-attention layer consists of several attention heads, and then the residual is computed by concatenating and projecting the attention heads output back to the image feature space $d_h$:
$$
\hat{\phi} = \phi + \textrm{Multi-Head-Attention}(\phi).
$$

\subsection{Style Aligned Image Set Generation}
\label{ourmethod}

The goal of our method is to generate a set of images $\mathcal{I}_1 \dots \mathcal{I}_n$ that are aligned with an input set of text prompts $y_1 \dots y_n$ and share a consistent style interpretation with each other. For example, see the garnered image set of toy objects in Fig.~\ref{fig:diagram} that are style-aligned with each other and to the input text on top.
A naïve way to generate a style aligned image set of different content is to use the same style description in the text prompts. As can be seen at the bottom of Fig.~\ref{fig:aligned}, generating different images using a shared style description of ``in minimal origami style'' results in an unaligned set, since each image is unaware of the exact appearance of other images in the set during the generation process.

The key insight underlying our approach is the utilization of the self-attention mechanism to allow communication among various generated images. This is achieved by sharing attention layers across the generated images.

Formally, let $Q_i$, $K_i$, and $V_i$ be the queries, keys, and values, projected from deep features $\phi_i$ of $\mathcal{I}_i$ in the set, then, the attention update for $\phi_i$ is given by:
\begin{equation}
    \textrm{Attention}(\queries_i, \keys_{\shared} , \values_{\shared}),
    \label{eq:shared-attention}
\end{equation}
where $\keys_{\shared} = \begin{bmatrix} K_1  \\ K_2  \\ \vdots \\ K_n \end{bmatrix}$ and $\values_{\shared}=\begin{bmatrix} V_1  \\ V_2  \\ \vdots \\ V_n  \end{bmatrix}$.\vspace{5pt}
 However, we have noticed that by enabling full attention sharing,  we may harm the quality of the generated set. As can be seen in Fig.~\ref{fig:ablation_qual} (bottom rows),  full attention sharing results in content leakage among the images. For example, the unicorns got green paint from the garnered dino in the set. Moreover, full attention sharing results with less diverse sets of the same set of prompts, see the two sets in Fig.~\ref{fig:ablation_qual} in bottom rows compared to the sets above.
 
To restrict the content leakage and allow diverse sets, we share the attention to only one image in the generated set (typically the first in the batch). That is,  \textit{\tarim{}} image features $\phi_t$ are attending to themselves and to the features of only one \textit{\refim{}} image in the set using Eq.~\ref{eq:shared-attention}.
As can be seen in Fig.~\ref{fig:ablation_qual} (middle), sharing the attention to only one image in the set results in diverse sets that share a similar style. However, in that case, we have noticed that the style of different images is not well aligned.
We suspect that this is due to low attention flow from the \refim{} to the \tarim{} image.

As illustrated in Fig. \ref{fig:diagram-right-side}, to enable balanced attention \refim{}, we normalize the queries $Q_t$ and keys $K_t$ of the target image using the queries $Q_r$ and keys $K_r$ of the \refim{} image using the adaptive normalization operation (AdaIN) \cite{huang2017arbitrary}:
$$
\hat{Q_t} = \textrm{AdaIN}(Q_t, Q_r) \;\;\;\;
\hat{K_t} = \textrm{AdaIN}(K_t, K_r),
$$
where the AdaIn operation is given by:
$$
\textrm{AdaIN}\left(x, y\right) = \sigma\left(y\right) \left(  \dfrac{x - \mu(x)}{\sigma(x)} \right) + \mu_y,
$$
and $\mu(x), \sigma(x) \in \reals^{d_k}$ are the mean and the standard deviation of queries and keys across different pixels. 
Finally, our shared attention is given by 
$$
\mathrm{Attention}(\hat{Q_t}, K_{rt}^T, V_{rt}),
$$
where $\keys_{rt} = \begin{bmatrix}  K_r  \\ \hat{K_t} \end{bmatrix}$ and $\values_{rt}=\begin{bmatrix} V_r  \\ V_t \end{bmatrix}$.

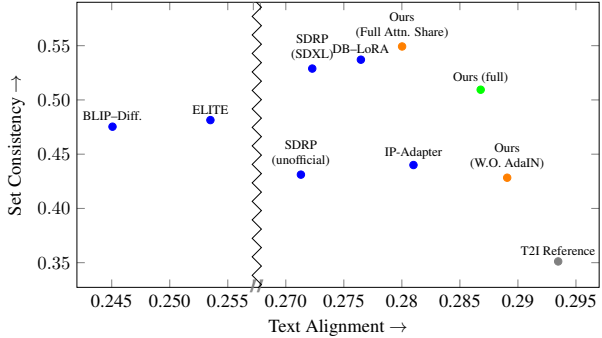
\begin{figure}[t]
        \begin{tikzpicture} [thick,scale=0.7, every node/.style={scale=1}]
       
            \def\MarkSize{.75em}
            \protected\def\ToWest#1{
              \llap{#1\kern\MarkSize}\phantom{#1}
            }
            \protected\def\ToSouth#1{
              \sbox0{#1}
              \smash{
                \rlap{
                  \kern-.5\dimexpr\wd0 + \MarkSize\relax
                  \lower\dimexpr.575em+\ht0\relax\copy0
                }
              }
              \hphantom{#1}
            }
            \begin{axis}[
                yticklabel=\pgfkeys{/pgf/number format/.cd,fixed,precision=2,zerofill}\pgfmathprintnumber{\tick},
                xticklabel=\pgfkeys{/pgf/number format/.cd,fixed,precision=3,zerofill}\pgfmathprintnumber{\tick},
                xlabel={Text Alignment $\rightarrow$}, 
                ylabel={Set Consistency $\rightarrow$},
                compat=newest,
                xmin=0.252,
                xmax=0.297,
                ymax=0.590,
                width=11.5cm,
                height=7cm,
                ytick={0.35, 0.40, 0.45, 0.50, 0.55}, 
                xticklabels=  {0.245, 0.250, 0.255, 0.270, 0.275, 0.28, 0.285, 0.29, 0.295, 0.30},
                xtick=        {0.255, 0.260, 0.265, 0.270, 0.275, 0.28, 0.285, 0.29, 0.295, 0.30}, 
                extra x ticks={0.2672,0.2678},
                extra x tick style={grid=none, tick label style={xshift=0cm,yshift=.280cm, rotate=0}},
                extra x tick label={\color{gray}{/\!\!/}}
            ]
                \addplot[
                    scatter/classes={a={blue}, b={red}, c={green}, o={orange}, g={gray}},
                    scatter,
                    mark=*, 
                    only marks, 
                    scatter src=explicit symbolic,
                    nodes near coords*={\Label},
                    visualization depends on={value \thisrow{label} \as \Label}
                ] table [meta=class] {
                    x y class label
                    0.29346888852119446 0.3511449992656708 g \scriptsize{T2I Reference}
                    0.27227795124053955 0.5288990139961243 a \scriptsize{\begin{tabular}{c}SDRP \\ (SDXL)\end{tabular} }
                    0.27130050349235535 0.4311416745185852 a \scriptsize{\begin{tabular}{c}SDRP \\ (unofficial)\end{tabular}}
                   0.2764682173728943 0.5370860695838928 a \scriptsize{DB--LoRA}
                    0.2867910225391388 0.5094115138053894 c \scriptsize{Ours (full)}
                    0.2890741229057312 0.4283123016357422  o \scriptsize{\begin{tabular}{c}Ours \\ (W.O. AdaIN)\end{tabular} }
                    0.28001922369003296 0.5492603778839111  o \scriptsize{\begin{tabular}{c}Ours \\ (Full Attn. Share)\end{tabular} } 
                    0.281 0.44 a \scriptsize{IP-Adapter}
                      0.26350343704223633 0.4814221262931824 a \scriptsize{ELITE}
                       0.2550660914182663  0.4753357768058777 a \scriptsize{BLIP--Diff.}
         
                };
                 \draw[black] decorate [decoration={zigzag}] {(axis cs:0.2675,0.3) -- (axis cs:0.2675,0.6)};
            
            \end{axis}

        \end{tikzpicture} 
    \caption{\textbf{Quantitative Comparison.} \emph{We compare the results of the different methods (blue marks) and our ablation experiments (orange marks)  in terms of text alignment (CLIP score) and set consistency (DINO embedding  similarity).}}
    \label{fig:quant}
\end{figure}

\begin{figure*}[h!]
\footnotesize
\centering
    \includegraphics[width=.98\textwidth]{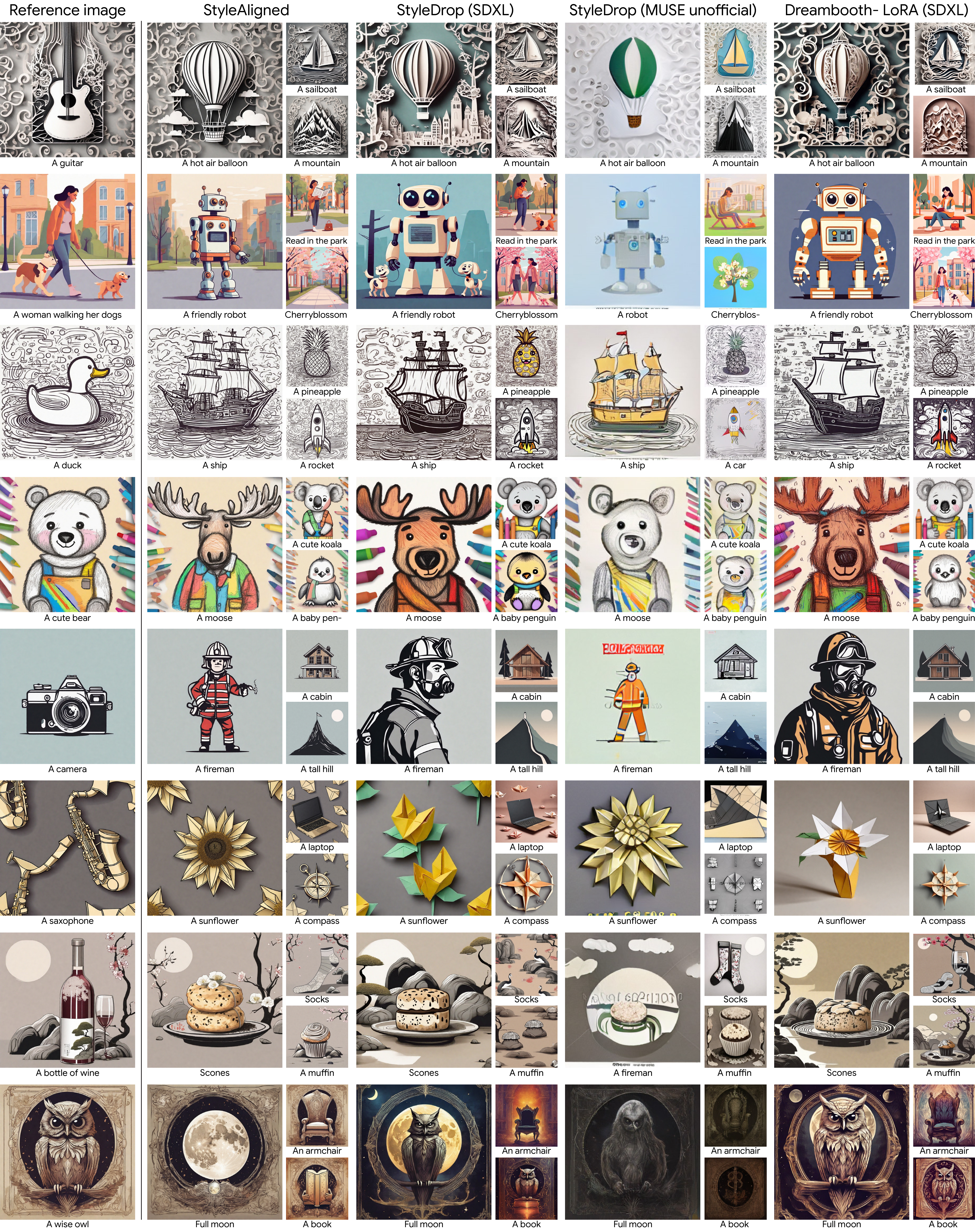}
\caption{\textbf{Qualitative comparison to personalization based methods.}}
\label{fig:comparisonqual}
\end{figure*}

\begin{table}[t!]
\caption{\textbf{ User evaluation for style aligned image set generation.} \textit{ In each question, the user was asked to select  between two image sets, Which is better in terms of style consistency and match to the text descriptions (see Sec.~\ref{sec:results}). We report the percentage of
judgments in favor of {\ourmethod} over 800 answers (2400 in total).}
}
 \scriptsize
 \centering
 \vspace{0.2cm}
\begin{tabular}{ccc}

\toprule
\begin{tabular}{c}StyleDrop \\ (unofficial MUSE)\end{tabular}&\begin{tabular}{c}StyleDrop \\ (SDXL) \end{tabular}&  \begin{tabular}{c}DreamBooth--LoRA \\ (SDXL) \end{tabular}  \\
\midrule
85.2 \% & 67.1 \% & 61.3\% \\
\bottomrule
\end{tabular}
\label{tab:user_study}
\end{table}

\section{Evaluations and Experiments}
\label{sec:results}
We have implemented our method over Stable Diffusion XL (SDXL) \cite{Podell2023SDXLIL} by applying our attention sharing overall $70$ self-attention layers of the model.
The generation of a four images set takes $29$ seconds  on a single $A100$ GPU. Notice that since the generation of the reference image is not influenced by other images in the batch, we can generate larger sets by fixing the prompt and seed of the reference image across the set generation.

For example, see the sets in Fig.~\ref{fig:aligned} and ~\ref{fig:diagram}.

\textbf{Evaluation set.}
With the support of ChatGPT,we have generated $100$ text prompts describing different image styles over four random objects.
For example, ``\{A guitar, A hot air balloon, A sailboat, A mountain\} in papercut art style.''
For each style and set of objects, we use our method to generate a set of images. The full list of prompts is provided in the appendix.

\textbf{Metrics.}
To verify that each generated image contains its specified object, we measure the CLIP cosine similarity  \cite{Radford2021LearningTV} between the image and 
the text description of the object.
In addition, we evaluate the style consistency of each generated set, by measuring the pairwise average cosine similarity between DINO VIT-B/8 \cite{caron2021emerging} embeddings of the generated images in each set. Following \cite{Ruiz2022DreamBoothFT, Voynov2023PET}, we used DINO embeddings instead of CLIP image embeddings for measuring image similarity, 
since CLIP was trained with class labels and therefore it might give a high score for different images in the set that have similar content but with a different style. On the other hand, DINO better distinguishes between different styles due to its self-supervised training. 

\subsection{Ablation Study}
The quantitative results are summarized in Fig.~\ref{fig:quant}, where the right--top place on the chart means better text similarity and style consistency, respectively. As a reference, we report the score obtained by generating the set of images using SDXL (T2I Reference) using the same seeds without any intervention. As can be seen, our method achieves a much higher style consistency score at the expense of text similarity. See qualitative comparison in Fig.~\ref{fig:aligned}.

In addition, we compared our method to additional two variants of the shared attention as described in Sec.~\ref{ourmethod}. The first variant uses full attention sharing (\textit{Full Attn. Share}) where the keys and values are shared between each pair of images in the set. In the second variant (\textit{W.A. AdaIN}) we omit the AdaIN operation over queries and keys. As expected, this \textit{Full Attn. Share} variant, results with higher style consistency and lower text alignment. As can be seen in Fig.~\ref{fig:ablation_qual}, \textit{Full Attn. Share} harms the overall quality of the image sets and diversity across sets. Moreover, our method without the use of AdaIN results in much lower style consistency. Qualitative results can be seen in Fig.~\ref{fig:ablation_qual}.  

\begin{figure}[t]
\footnotesize
\centering
    \includegraphics[width=1\columnwidth]{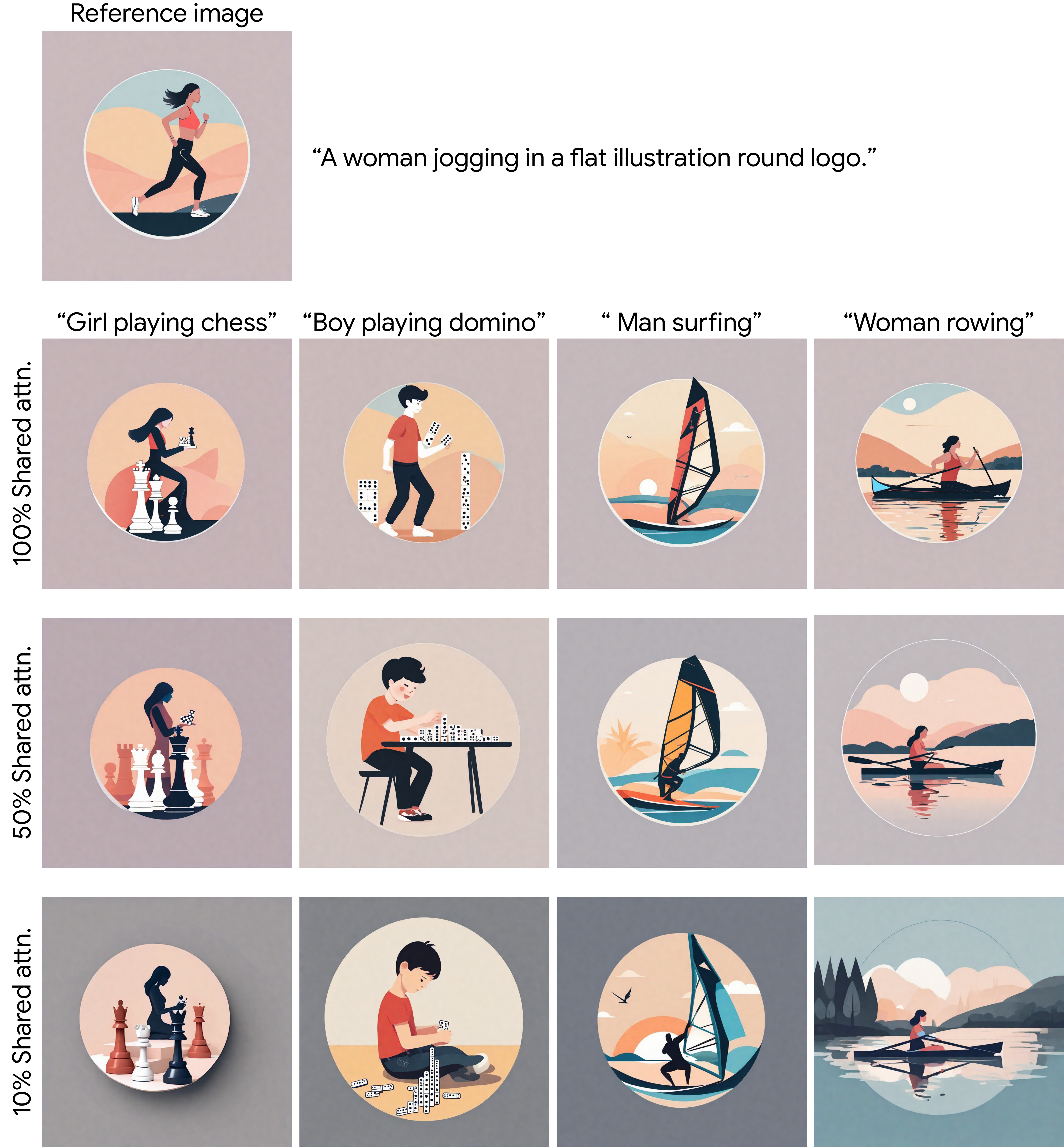}
\caption{\textbf{Varying level of attention sharing.} \emph{By reducing the number of shared attention layers, i.e., allowing only self-attention in part of the layers, we can get more varied results (bottom rows) at the expense of style alignment (top row). 
}
}
\vspace{-3mm}
\label{fig:varied-attnetion}
\end{figure}

\begin{figure}[b]
\footnotesize
\centering
\vspace{-3mm}
    \includegraphics[width=1\columnwidth]{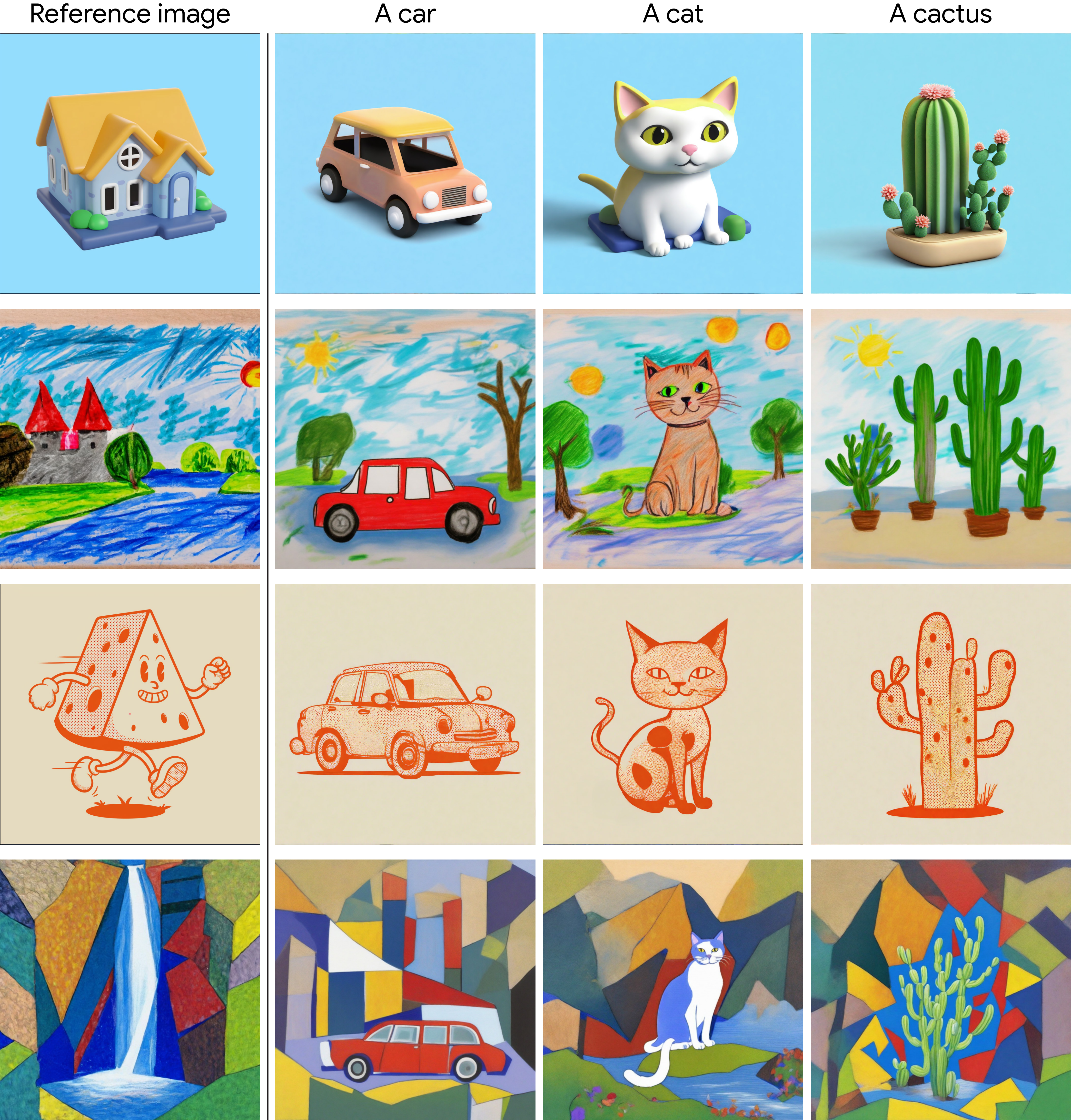}
\caption{\textbf{Style aligned image generation to an input image.} \emph{Given an input reference image (left column) and text description, we first apply DDIM inversion over the image to get the inverted diffusion trajectory $x_T, x_{T-1} \dots x_0$. Then, starting from $x_T$ and a new set of prompts, we apply our method to generate new content (right columns) with an aligned style to the input. }}
\label{fig:reals}
\end{figure}

\subsection{Comparisons}
For baselines, we  compare our method to T2I personalization methods.
We trained StyleDrop~\cite{sohn2023styledrop} and DreamBooth~\cite{Ruiz2022DreamBoothFT} over the first image in each set of our  evaluation data, and use the trained personalized weights to generate the additional three images in each set.
We use a public unofficial implementation of StyleDrop~\footnote[1]{\url{github.com/aim-uofa/StyleDrop-PyTorch}} (SDRP--unofficial) over non-regressive T2I model. Due to the large quality gap between the unofficial MUSE model~\footnote[2]{\url{github.com/baaivision/MUSE-Pytorch}} to the official one~\cite{chang2023muse}, we follow StyleDrop and implement an adapter model over SDXL (SDRP--SDXL), where we train a low rank linear layer after each Feed-Forward layer at the model's attention blocks.
For training DreamBooth, we adapt the LoRA~\cite{lora, lora_diffusion} variant (DB--LoRA) over SDXL using the public huggingface--diffusers implementation~\footnote[3]{\url{github.com/huggingface/diffusers}}. We follow the hyperparameters tuning reported in~\cite{sohn2023styledrop} and train both SDRP--SDXL and DB--LoRA for 400 steps to prevent overfitting to the style training image.

As can be seen in the qualitative comparison, Fig.~\ref{fig:comparisonqual}, the image sets generated by our method, are more consistent across style attributes like color palette, drawing style, composition, and pose. Moreover, the personalization-based methods may leak the content of the training reference image (on the left) when generating the new images. For example, see the repeated woman and dogs in the results of DB--LoRA and SDRP--SDXL at the second row or the repeated owl at the bottom row. Similarly, because of the content leakage, these methods obtained lower text similarity scores and higher set consistency scores compared to our method. 

We also apply two encoder-based personalization methods ELITE~\cite{ELITE-implmentation},  IP--Adapter~\cite{Ye2023IPAdapterTC}, and BLIP--Diffusion~\cite{Li2023BLIPDiffusionPS} over our evaluation set. These methods receive as input the first image in each set and use its embeddings to generate images with other content. Unlike the optimization-based techniques, these methods operate in a much faster feed-forward diffusion loop, like our method. However,
as can be seen in Fig.~\ref{fig:quant}, their performance for style aligned image generation is poor compared to the other baselines. We argue that current  encoder-based personalization techniques struggle to disentangle the content and the style of the input image. We supply qualitative results in appendix \ref{appen:compare}.

\textbf{User Study.}
In addition to the automatic evaluation, we conducted a user study over the results of our method, StyleDrop (unofficial MUSE), StyleDrop (SDXL), and DreamBooth--LoRA (SDXL). In each question, we randomly sample one of the evaluation examples and show the user the 4 image set that resulted from our and another method (in a random order). The user had to choose which set is better in terms of style consistency, and text alignment. A print screen of the user study format is provided in the appendix. Overall, we collected 2400 answers from 100 users using the Amazon Mechanical Turk service. The results are summarized in Tab.~\ref{tab:user_study} where for each method, we report the  percentage of judgments in our favor. As can be seen,  most participants favored our method by a large margin. More information about our user study can be found in appendix \ref{sec:user}.

\subsection{Additional Results}

 \textbf{Style Alignment Control.} We provide means of control over the style alignment to the reference image by applying the shared attention over only part of the self-attention layers. As can be seen in Fig.~\ref{fig:varied-attnetion}, reducing the number of shared attention layers results with a more diverse image set, which still shares common attributes with the reference image.

\textbf{\ourmethod{} from an Input Image.} 
To generate style-aligned images to an input image, we apply DDIM inversion \cite{song2020denoising} using a provided text caption. Then, we apply our method to generate new images in the style of the input using the inverted diffusion trajectory $x_T, x_{T-1}, \dots x_0$ for the reference image.
Examples are shown in Fig.~\ref{fig:reals} ,\ref{fig:babylon}, where we use BLIP captioning \cite{blipcap} to get a caption for each input image. For example, we used the prompt ``A render of a house with a yellow roof'' for the DDIM inversion of the top example and replaced the word house with other objects to generate the style-aligned images of a car, a cat, and a cactus.
Notice that this method does not require any optimization. However, DDIM inversion may fail \cite{mokady2022null} or results with an erroneous trajectory \cite{HubermanSpiegelglas2023}. More results and analysis, are provided in  appendix~\ref{appen:results}

\begin{figure}
    \centering
    \includegraphics[width=\columnwidth]{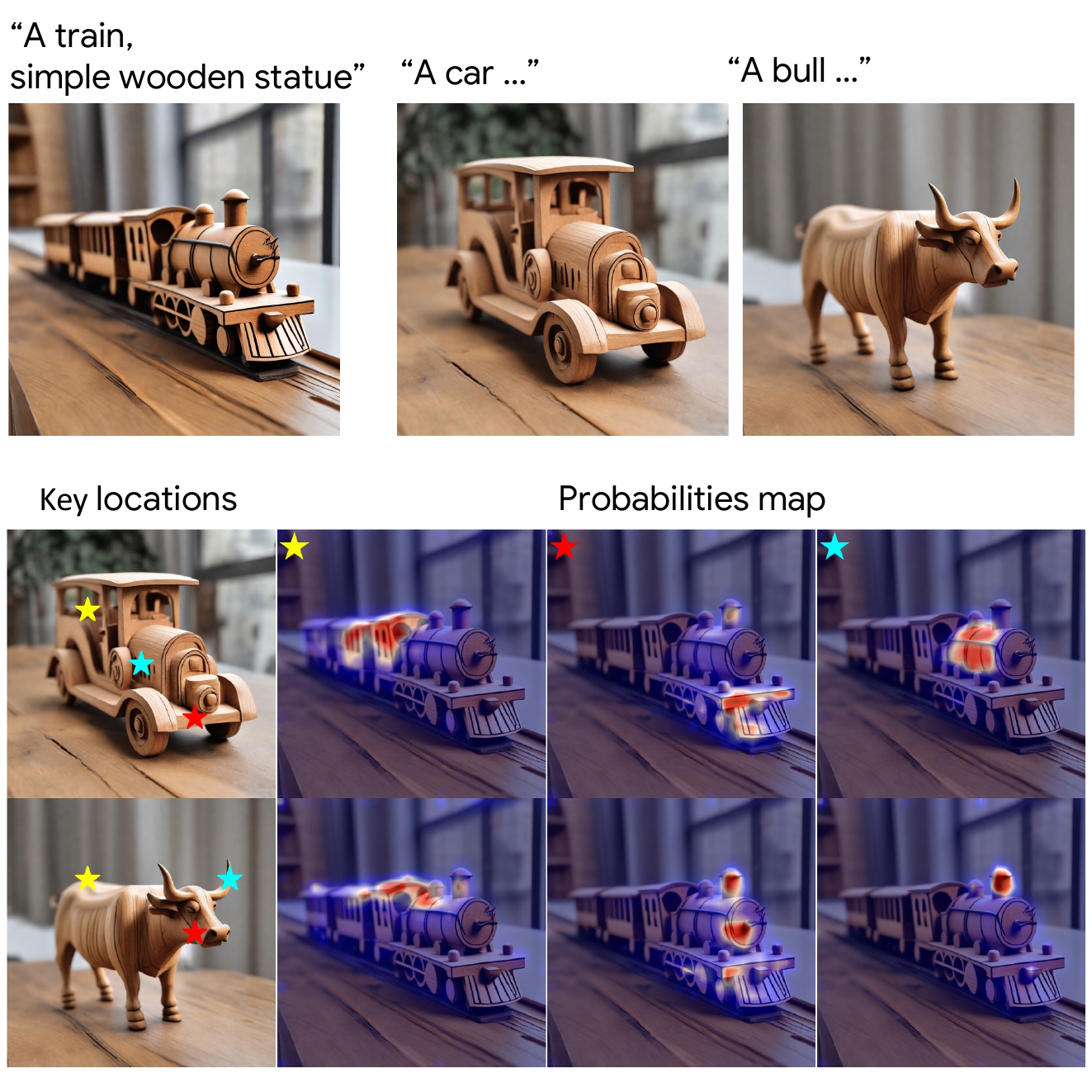}
    \caption{\textit{Self-Attention probabilities maps from different generated image locations (\textbf{Key locations} column) to the reference train image with the target style (top-left).}}
    \label{fig:sa-vis}
\end{figure}

\begin{figure}[t]
\footnotesize
\centering
    \includegraphics[width=1\columnwidth]{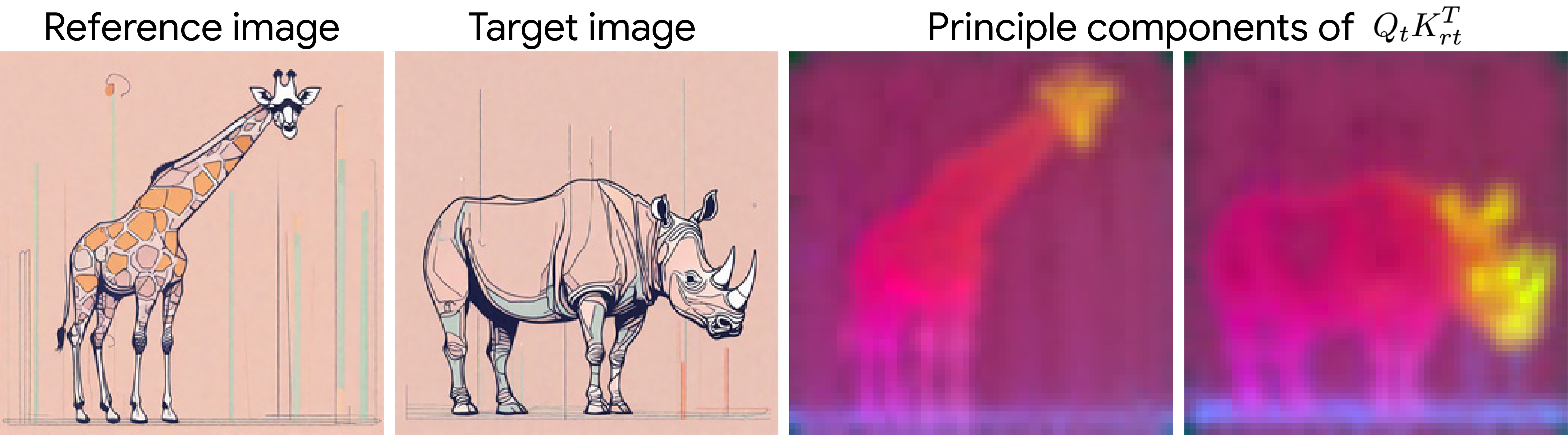}
\caption{\textbf{Principle components of the shared attention map.} \textit{On right, we visualize the principle components of the shared attention map between the reference giraffe and the target rhino generated images. The three largest components of the shared maps are encoded in RGB channels.}}
\vspace{-3mm}
\label{fig:pca-rhino}
\end{figure}

\textbf{Shared Self-Attention Visualization.}
Figure \ref{fig:sa-vis} depicts the self-attention probabilities from a generated target image to the reference style image. In each of the rows, we pick a point on the image and depict the associated probabilities map for the token at this particular point. Notably probabilities mapped on the reference  image are semantically close to the query point location. This suggests that the self-attention tokens sharing do not perform a global style transfer, but rather match the styles in a semantically meaningful way \cite{alaluf2023crossimage}.
In addition, Figure \ref{fig:pca-rhino} visualizes the three largest components of the average shared attention maps of the rhino image, encoded in RGB channels. Note that the shared attention map is composed of both self-attention 
and cross-image attention to the giraffe. As can be seen, the components highlight semantically related regions like the bodies, heads, and the background in the images.

\begin{figure}[t]
\footnotesize
\centering
    \includegraphics[width=1\columnwidth]{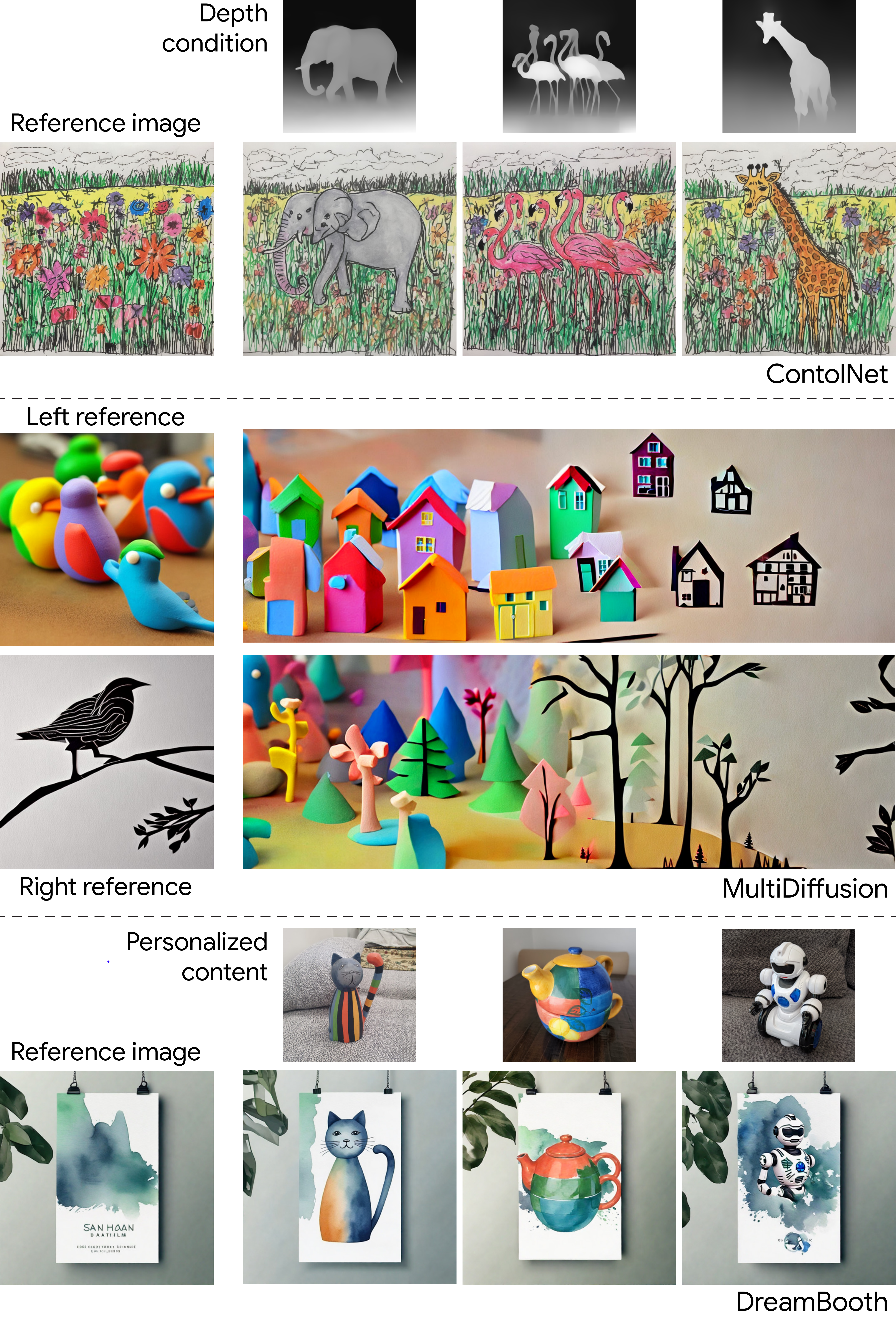}
\caption{\textbf{StyleAligned with other methods.} \emph{On top, StyleAligned is combined with ControlNet to generate style-aligned images conditioned on depth maps. In the middle, our method combined with MultiDiffusion to generate panorama images that share multiple styles. On the bottom, style consistent and personalized content created by combining our method with pre-trained personalized DreamBooth--LoRA models.}}
\label{fig:with_other_methods}
\end{figure}

\textbf{StyleAligned with Other Methods.}
Since our method doesn't require training or optimization, it can be easily combined on top of other diffusion based methods to generate style-consistent image sets.
Fig.~\ref{fig:with_other_methods} shows several such examples where we combine our method with ControlNet \cite{zhang2023controlnet}, DreamBooth~\cite{ruiz2023hyperdreambooth} and MultiDiffusion \cite{BarTal2023MultiDiffusionFD}. 
More examples and details about the integration of StyleAligned with other methods can be found in appendix \ref{sec:integration}.

\begin{figure*}[h!]
    \centering
    \includegraphics[width=\textwidth]{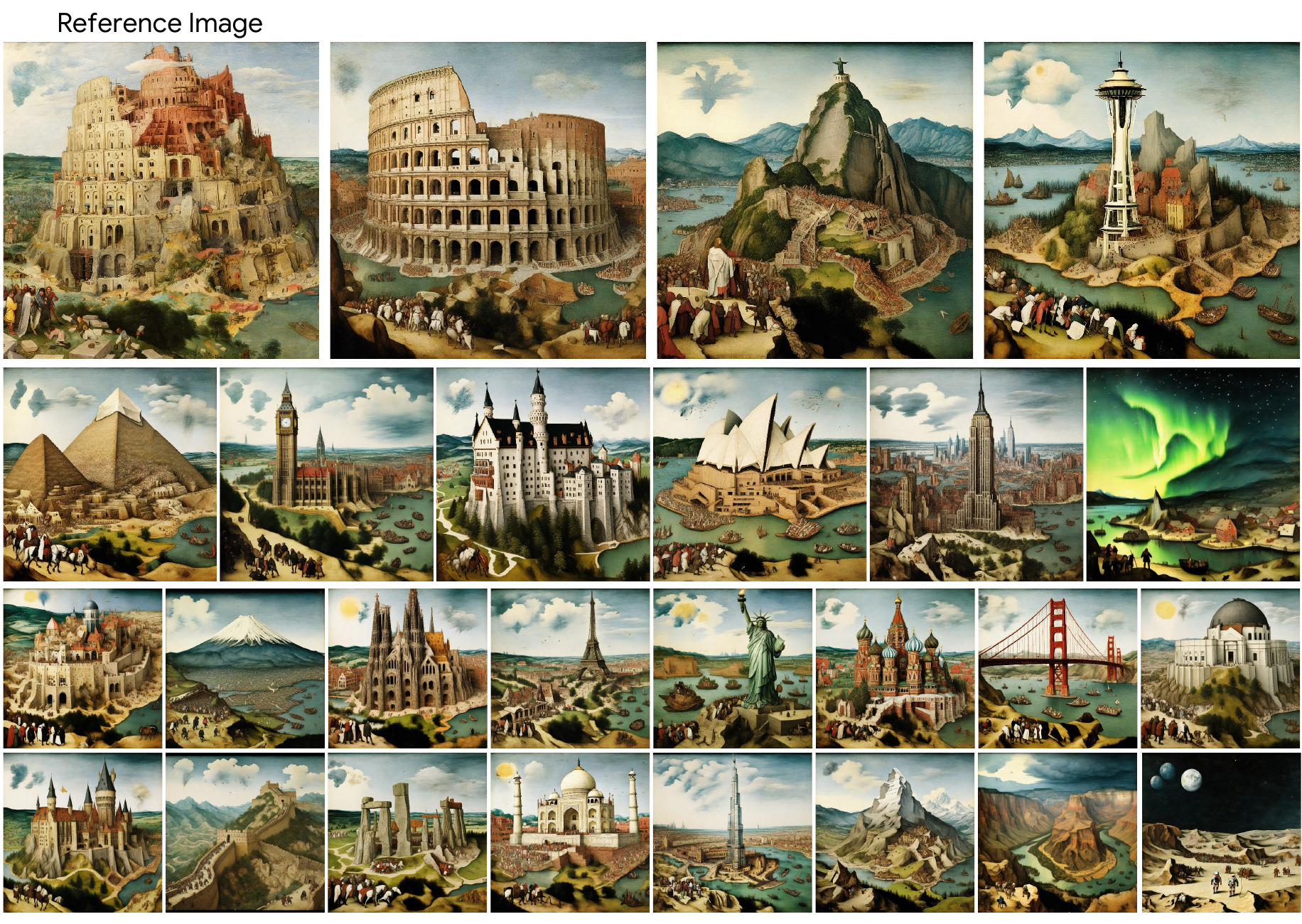}
    \caption{\textbf{Various remarkable places depicted with the style taken from Bruegels' \textit{``The Tower of Babel''}.} \\
     \textit{Top row: Rome Colosseum, Rio de Janeiro, Seattle Space Needle.}
     }
    \label{fig:babylon}
\end{figure*}

\section{Conclusions}

We have presented \ourmethod{}, which addresses the challenge of achieving style-aligned image generation within the realm of large-scale Text-to-Image models. By introducing minimal attention sharing operations with AdaIN modulation during the diffusion process, our method successfully establishes style-consistency and visual coherence across generated images. The demonstrated efficacy of \ourmethod{} in producing high-quality, style-consistent images across diverse styles and textual prompts underscores its potential in creative domains and practical applications. Our results affirm \ourmethod{} capability to faithfully adhere to provided descriptions and reference styles while maintaining impressive synthesis quality.

In the future we would like to explore the scalability and adaptability of \ourmethod{} to have more control over the shape and appearance similarity among the generated images.
Additionally, due to the limitation of current diffusion inversion methods, a promising direction is to leverage \ourmethod{} to assemble a style-aligned dataset which then can be used to train style condition text-to-image models.

\section{Acknowledgement}
We thank Or Patashnik, Matan Cohen, Yael Pritch, and Yael Vinker for their valuable inputs that helped improve this work.

{
    \small
    \bibliographystyle{ieee_fullname}
    \bibliography{egbib}
}

\appendix

\section*{\Large{Appendix}}

\section{StyleAligned from an Input Image}
\label{appen:results}

\begin{figure}[t]
    \centering
    \includegraphics[width=\columnwidth]{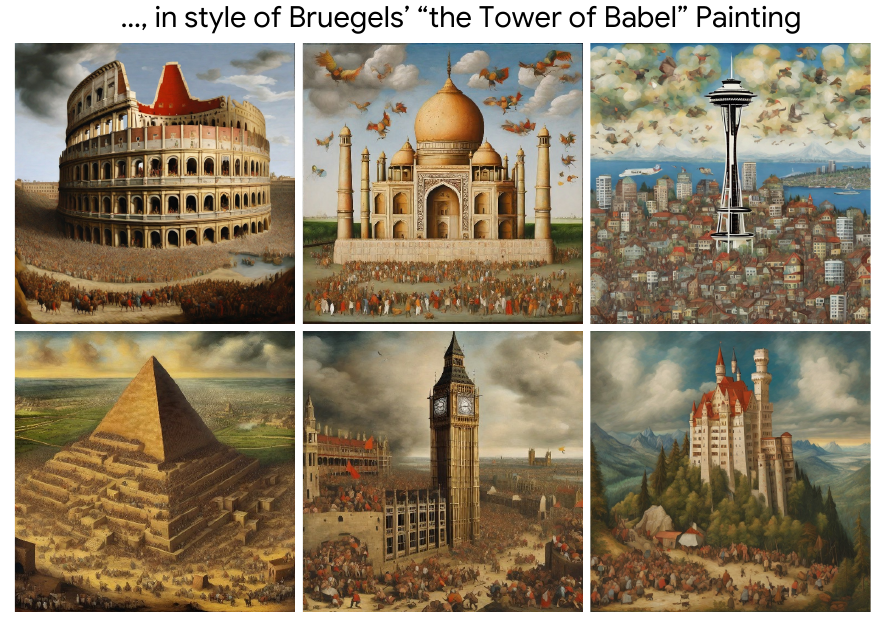}
    \vspace{-0.5cm}
    \caption{\textbf{Text-to-image generation with explicit style description.}\textit{ Unlike our approach, this fails to produce fine and style-aligned results. See Fig. \ref{fig:babylon} to inspect our method results.}}
    \label{fig:babylon_baseline}
    \vspace{-.4cm}
\end{figure}

Figure \ref{fig:babylon} shows our techniques being applied for style transfer for the Peter Bruegels' "The Tower of Babel" to multiple places around the world. As for the prompt we always use the places' followed by \textit{"Pieter Bruegel Painting"}, e.g.``\textit{Rome Coliseum, Pieter Bruegel Painting}''. Even though the original masterpiece is known to model, it fails to reproduce its style with only text guidance. Fig. \ref{fig:babylon_baseline} shows some of the places generated with the direct instruction to resemble the original painting, without self-attention sharing. Notably, the model fails to produce an accurate style alignment with the original picture.

Further examples of style transferring from real examples are presented  in Figures \ref{fig:real_grid_1} and \ref{fig:real_grid_2}.

\begin{figure}[t!]
    \centering
    \includegraphics[width=\columnwidth]{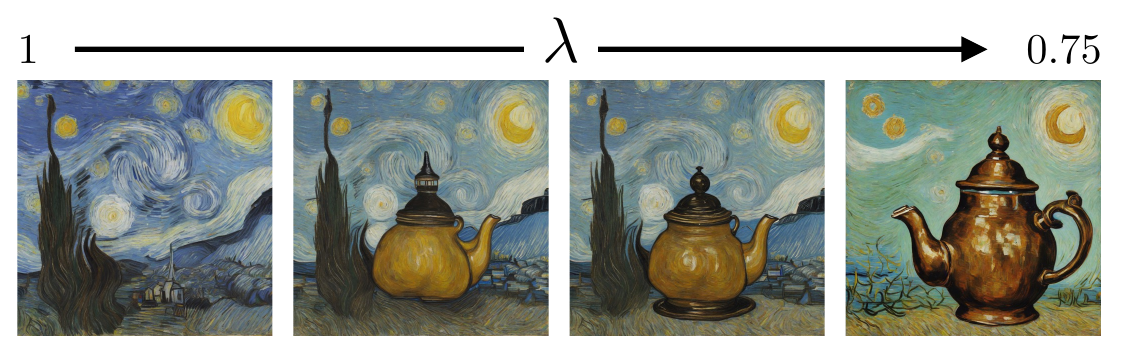}
    \vspace{-0.6cm}
    \caption{\textbf{Reference attention rescaling factor variation used for extremely popular reference image assets.}}
    \vspace{-.4cm}
    \label{fig:rescale}
\end{figure}

We also noticed that once the style transfer is performed from an extremely famous image, the default approach may sometimes completely ignore the target prompt, generating an image almost identical to the reference. We suppose that this happens because the outputs of the denoising model for the famous reference image have very high confidence and activations magnitudes. Thus in the shared self-attention, most of the attention is taken by the reference keys. To compensate for it, we propose the simple trick of the attention scores rescaling. In the self-attention sharing mechanism, for some fixed scale $\lambda < 1$, we rescale the queries and keys products conducting the new scores $\lambda \cdot \left<Q, K_{\mathrm{target}}\right>$. We apply this only to the reference image keys. First, this suppresses extra-high keys. Also, this makes the attention scores more uniformly distributed, encouraging the generated image to capture style aggregated from the whole reference image. Fig.~\ref{fig:rescale} demonstrates the rescaling factor variation effect for the particularly popular reference \textit{"Starr Night"} by Van Gogh. Notably, without rescaling, the model generates an image almost identical to the reference, while the scale relaxation produces a plausible transfer.

\section{Integration with Other Methods}
\label{sec:integration}
 Below, we show different examples where our method can provide style aligned image generation capability on top of different  diffusion-based image generation methods.

\textbf{Style Aligned Subject Driven Generation.}
To use our method on top of a personalized diffusion model, first, given a collection of images (3-6) of the personalized content, we follow DreamBooth--LoRA training ~\cite{ruiz2023hyperdreambooth, lora}  where the layers of the attention layers are fine-tuned via low-rank adaptation weights (LoRA).
Then, during inference, we apply our method by sharing the attention of personalized generated images with a generated reference style image. During this process, the LoRA weights are used only for the generation of personalized content.
Examples of style aligned personalized images are shown in Fig.~
The results of our method on top of different personalized models are shown in Fig.~\ref{fig:app-db} where in each column we fine-tuned the SDXL model over the image collection on top and generated the personalized  content with the reference images on the left.
It can be seen that in some cases, like in the backpack photos on the right, the subject in the image remains in the same style as in the original photos. This is a known limitation of training-based personalization methods ~\cite{Tewel2023KeyLockedRO} which we believe can be improved by applying our method over other T2I personalization techniques \cite{Kumari2022MultiConceptCO, Han2023SVDiffCP} or more careful search for training hyperparameters  that allow better generalization of the personalized model to different styles.

\textbf{Style Aligned MultiDiffusion Image Generation.}
Bar et al. \cite{BarTal2023MultiDiffusionFD} presented MultiDiffusion, a method for generating images in any resolution by aggregating diffusion predictions of overlapped squared crops. Our method can be used on top of MultiDiffusion by enabling our shared attention between the crops to a reference image that is generated in parallel.
Fig.~\ref{fig:app-multi} shows style aligned panorama images generated with MultiDiffusion in conjunction with our method using the public implementation of MultiDiffusion over Stable Diffusion V2~\cite{panorama-pipe}.
Notice that compared to a \textit{vanilla} MultiDiffusion image generation (small images in~\ref{fig:app-multi}), our method not only enables the generation of style aligned panoramas but also helps to preserve the style within each image.

\textbf{StyleAligned with Additional Conditions.}
Lastly, we show how our method can be combined with ContolNet \cite{zhang2023controlnet} which enriches the conditioning signals of diffusion text-to-image generation to include additional inputs, like depth map and pose.
ContolNet injects the additional information by predicting residual features that are added to the diffusion image features outputs of  the down and middle U-Net blocks.
Similar to previous modifications, we apply StyleAligned image generation by sharing the attention of ControlNet conditioned images to a reference image that isn't conditioned on additional input.
Fig.~\ref{fig:app-depth} shows style aligned image set (different rows) that are conditioned on depth maps
(different columns) using ControlNet depth encoder over SDXL~\cite{controlnet-depth}.
Fig.~\ref{fig:app-pose} shows style aligned image set (different rows) that are conditioned on pose estimation obtained by OpenPose \cite{openpose} (different columns) using ControlNet pose encoder over SDXL~\cite{controlnet-pose}.

\section{Additional Comparisons}
\label{appen:compare}
We provide additional comparisons of our method to encoder-based text-to-image personalization methods and editing approaches over the evaluation set presented in Section 4 in the main paper. Table \ref{tab:quant} summarized the full quantitative results presented in the paper and here.

\textbf{Encoder Based Approaches}
As reported in the paper, we compare our method to encoder-based text-to-image personalization methods: BLIP-Diffusion \cite{Li2023BLIPDiffusionPS}, ELITE \cite{ELITE-implmentation}, and IP-Adapter \cite{Ye2023IPAdapterTC}.
These methods train an image encoder and fine-tune the T2I diffusion model to be conditioned on visual input. 
Fig.~\ref{fig:comparison-encoders} shows a qualitative comparison on the same set shown in the paper (Fig. 7). As can be seen, our image sets are more consistent and aligned to the reference. Notice that, currently, only IP-Adapter provides an encoder model for Stable Diffusion XL (SDXL). Nevertheless, BLIP-Diffusion and ELITE struggle to produce consistent image sets that match the text descriptions.

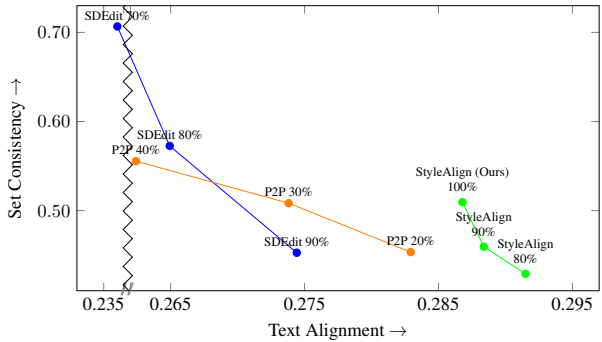
\begin{figure}[h]
        \begin{tikzpicture} [thick,scale=0.7, every node/.style={scale=1}]
            \def\MarkSize{.75em}
            \protected\def\ToWest#1{
              \llap{#1\kern\MarkSize}\phantom{#1}
            }
            \protected\def\ToSouth#1{
              \sbox0{#1}
              \smash{
                \rlap{
                  \kern-.5\dimexpr\wd0 + \MarkSize\relax
                  \lower\dimexpr.575em+\ht0\relax\copy0
                }
              }
              \hphantom{#1}
            }
            \begin{axis}[
                yticklabel=\pgfkeys{/pgf/number format/.cd,fixed,precision=2,zerofill}\pgfmathprintnumber{\tick},
                xticklabel=\pgfkeys{/pgf/number format/.cd,fixed,precision=2,zerofill}\pgfmathprintnumber{\tick},
                xlabel={Text Alignment $\rightarrow$}, 
                ylabel={Set Consistency $\rightarrow$},
                compat=newest,
                xmin=0.258,
                xmax=0.297,
                ymin=0.41,
                ymax=0.730,
                width=11.5cm,
                height=7cm,
                ytick={ 0.40, 0.50, 0.60, 0.70, 0.80}, 
                xticklabels={0.235, 0.265, 0.275, 0.285, 0.295}, 
                xtick={0.26, 0.265, 0.275, 0.285, 0.295}, 
                extra x ticks={0.2615,0.262},
                extra x tick style={grid=none, tick label style={xshift=0cm,yshift=.280cm, rotate=0}},
                extra x tick label={\color{gray}{/\!\!/}}
            ]
                \addplot[
                    scatter/classes={a={blue}, b={red}, c={green}, o={orange}, g={gray}},
                    scatter,
                    mark=*, 
                    color=blue,
                    scatter src=explicit symbolic,
                    nodes near coords*={\Label},
                    visualization depends on={value \thisrow{label} \as \Label}
                ] table [meta=class] {
                    x y class label
                  
                     0.26102014780044556 0.706694483757019 a \scriptsize{\color{black}{SDEdit 70\%}}
                      0.2649304270744324  0.5725564956665039 a \scriptsize{\color{black}SDEdit 80\%}
                       0.2744183201789856 0.4526869761943817 a \scriptsize{\color{black}SDEdit 90\%}
    
                };
                \addplot[
                    scatter/classes={a={blue}, b={red}, c={green}, o={orange}, g={gray}},
                    scatter,
                    mark=*, 
                    color=orange,
                    scatter src=explicit symbolic,
                    nodes near coords*={\Label},
                    visualization depends on={value \thisrow{label} \as \Label}
                ] table [meta=class] {
                    x y class label
                   
                     0.2624060044288635 0.5555180778503418 o \scriptsize{\color{black}{P2P 40\%}}
                      0.2738121449947357  0.5082688331604004 o \scriptsize{\color{black}P2P 30\%}
                       0.2829250991344452 0.45338791608810425 o \scriptsize{\color{black}P2P 20\%}
    
                };
                \addplot[
                    scatter/classes={a={blue}, b={red}, c={green}, o={orange}, g={gray}},
                    scatter,
                    mark=*, 
                    color=green,
                    scatter src=explicit symbolic,
                    nodes near coords*={\Label},
                    visualization depends on={value \thisrow{label} \as \Label}
                ] table [meta=class] {
                    x y class label
                    0.2867910225391388 0.5094115138053894  c \scriptsize{\color{black}{\begin{tabular}{c} StyleAlign (Ours)\\ 100\% \end{tabular}}}
                    0.28838321566581726 0.4594515264034271 c \scriptsize{\color{black}{\begin{tabular}{c} StyleAlign  \\ 90\%\end{tabular}}}
                    0.29149970412254333 0.4289761483669281 c \scriptsize{\color{black}{\begin{tabular}{c} StyleAlign  \\ 80\%\end{tabular}}}
                };
                    \addplot[
                    scatter/classes={a={blue}, b={red}, c={green}, o={orange}, g={gray}},
                    scatter,
                    mark=*, 
                    only marks, 
                    scatter src=explicit symbolic,
                    nodes near coords*={\Label},
                    visualization depends on={value \thisrow{label} \as \Label}
                ] table [meta=class] {
                    x y class label
                   0.29346888852119446 0.3511449992656708 g \scriptsize{T2I Reference}
                };

                \draw[black] decorate [decoration={zigzag}] {(axis cs:0.2618,0.1) -- (axis cs:0.2618,0.9)};
            
            \end{axis}
            
        \end{tikzpicture} 
    \caption{\textbf{Quantitative Comparison to zero shot editing approaces.} \emph{We compare the results of the different methods in terms of text alignment (CLIP score) and set consistency (DINO embedding  similarity).}}
    \vspace{-0.4cm}
    \label{fig:quant-zero}
\end{figure}

\begin{table}[t!]
 \caption{\textbf{Full quantitative  comparison for style aligned image generation.} We evaluate the generated image sets in terms of 
of text alignment (CLIP score) and set consistency (DINO embedding similarity). $\pm X$ denotes the standard deviation of the score across 100 image set results.}
 \footnotesize
 \centering
\begin{tabular}{l @{\hskip .5\tabcolsep}  c @{\hskip .5\tabcolsep} c @{\hskip .5\tabcolsep} }

\toprule
 Method & \begin{tabular}{c}Text Alignment \\ (CLIP $\uparrow$) \end{tabular}& \begin{tabular}{c}Set Consistency \\ (DINO $\uparrow$) \end{tabular} \\
StyleDrop (SDXL) & $0.272 \pm 0.04$  & $0.529 \pm 0.15$  \\ 
StyleDrop (unofficial MUSE) & $0.271 \pm 0.04$ & $0.301 \pm 0.14$ \\ 
DreamBooth-LoRA (SDXL)  & $0.276 \pm 0.03$ & $0.537 \pm 0.17$ \\ 
IP-Adapter (SDXL)  & $0.281 \pm 0.03$ & $0.44 \pm 0.13$ \\ 
ELITE (SD 1.4)  & $0.253 \pm 0.03$ & $0.481 \pm 0.13$ \\ 
BLIP-Diffusion (SD 1.4)  & $0.245 \pm 0.04$ & $0.475 \pm 0.12$ \\ 
Prompt-to-Prompt (SDXL)  & $0.283 \pm 0.03$ & $0.454 \pm 0.18$ \\ 
SDEdit (SDXL)  & $0.274 \pm 0.03$ & $0.453 \pm 0.16$ \\ 
\midrule
StyleAligned (SDXL) & $0.287 \pm 0.03$ & $0.51 \pm 0.14$ \\ 
StyleAligned (W.O. AdaIN) & $0.289 \pm 0.03$ &  $0.428 \pm 0.14$ \\ 
StyleAligned (Full Attn.) & $0.28 \pm 0.03$ & $0.55 \pm 0.15$ \\ 

\bottomrule
\end{tabular}
\vspace{-.3cm}
\label{tab:quant}
 \end{table}

\textbf{Zero Shot Editing Approaches}
Other baselines that can be used for style aligned image set generation are diffusion-based editing methods when applied over the reference images. However, unlike our method, these methods assume structure preservation of the input image.
We report the results of two diffusion-based editing approaches: SDEdit~\cite{meng2021sdedit} and Prompt-to-Prompt (P2P) \cite{Hertz2022PrompttoPromptIE} in Fig.~\ref{fig:quant-zero}. Notice that similar to our method, these methods provide a level of control 
that trade-off between alignment to text and alignment to the input image. To get higher text alignment, SDEdit can be applied over an increased percentage of diffusion steps, and P2P can reduce the number of attention injection steps. Our method can achieve higher text alignment, as described in Section 4 in the main paper, by using our shared  attention over only  a subset of self-attention layers. Fig.~\ref{fig:quant-zero} presents the trade-off of the results over the different methods.
As can be seen, only our method can achieve text alignment while preserving high set consistency.

\section{User Study and Evaluation Settings}
\label{sec:user}
As described in the main paper, we generate the images for evaluation using a list of 100 text prompts where each prompt describes 4 objects in the same style. The full list is provided at the end of supplementary materials~\ref{fig:list-prompts}.
We evaluated the results of the different methods using the automatic CLIP and DINO scores and through user evaluation. The format of the user study is provided in Fig.~\ref {fig:screenshot} where the user has to select between the results of two methods. For each method from StyleDrop (SDXL), StyleDrop (unofficial Muse), and DreamBooth-LoRA (SDXL), we collected 800 answers compared to our results. In total, we collected 2400 answers from 100 participants.

\begin{figure*}[t]
    \centering
    \includegraphics[width=\textwidth]{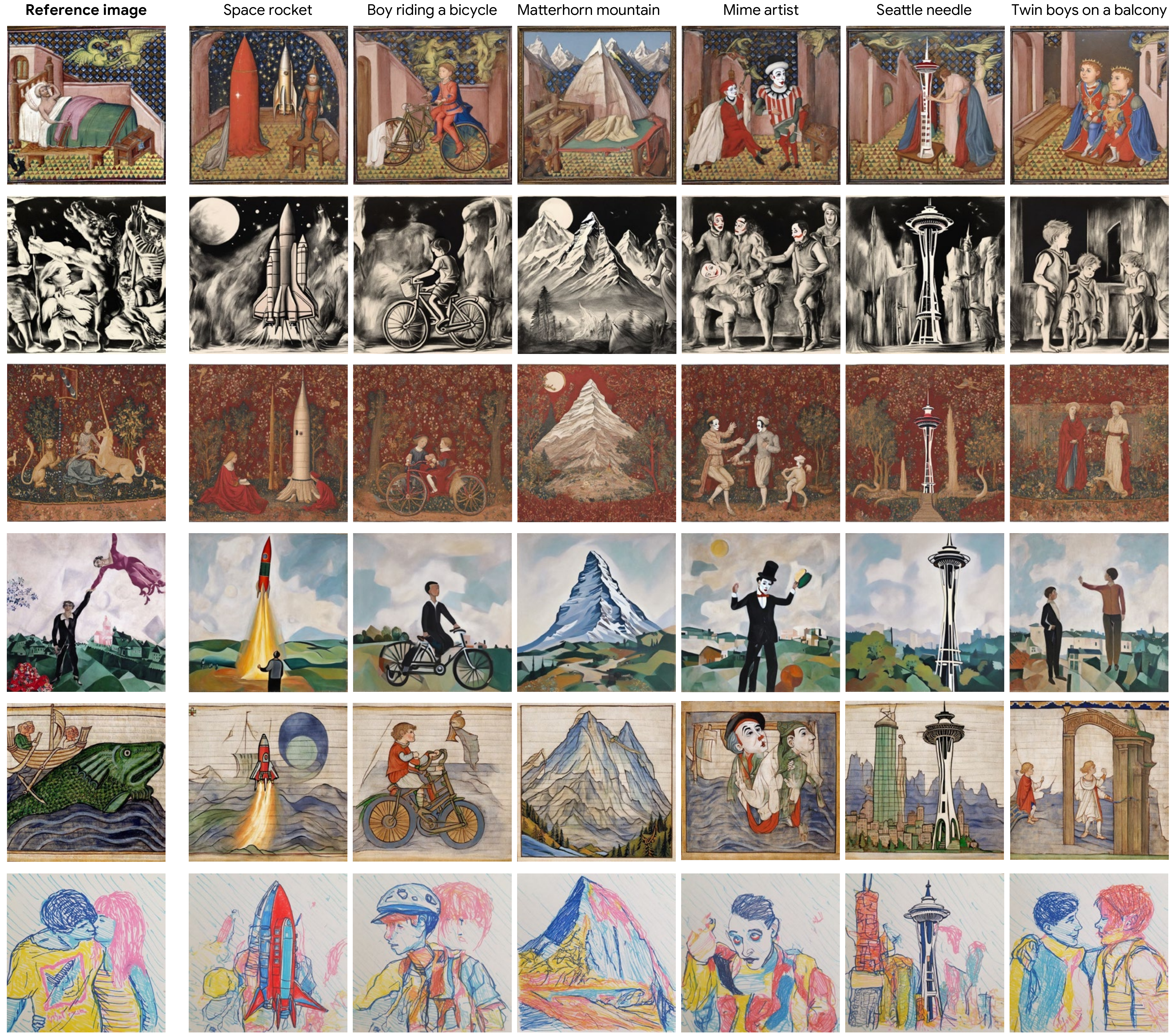}
    \caption{\textbf{Samples of the proposed style transfer techniques applied for a variety of different images and target prompts.}}
    \label{fig:real_grid_1}
\end{figure*}
\begin{figure*}[t]
    \centering
    \includegraphics[width=\textwidth]{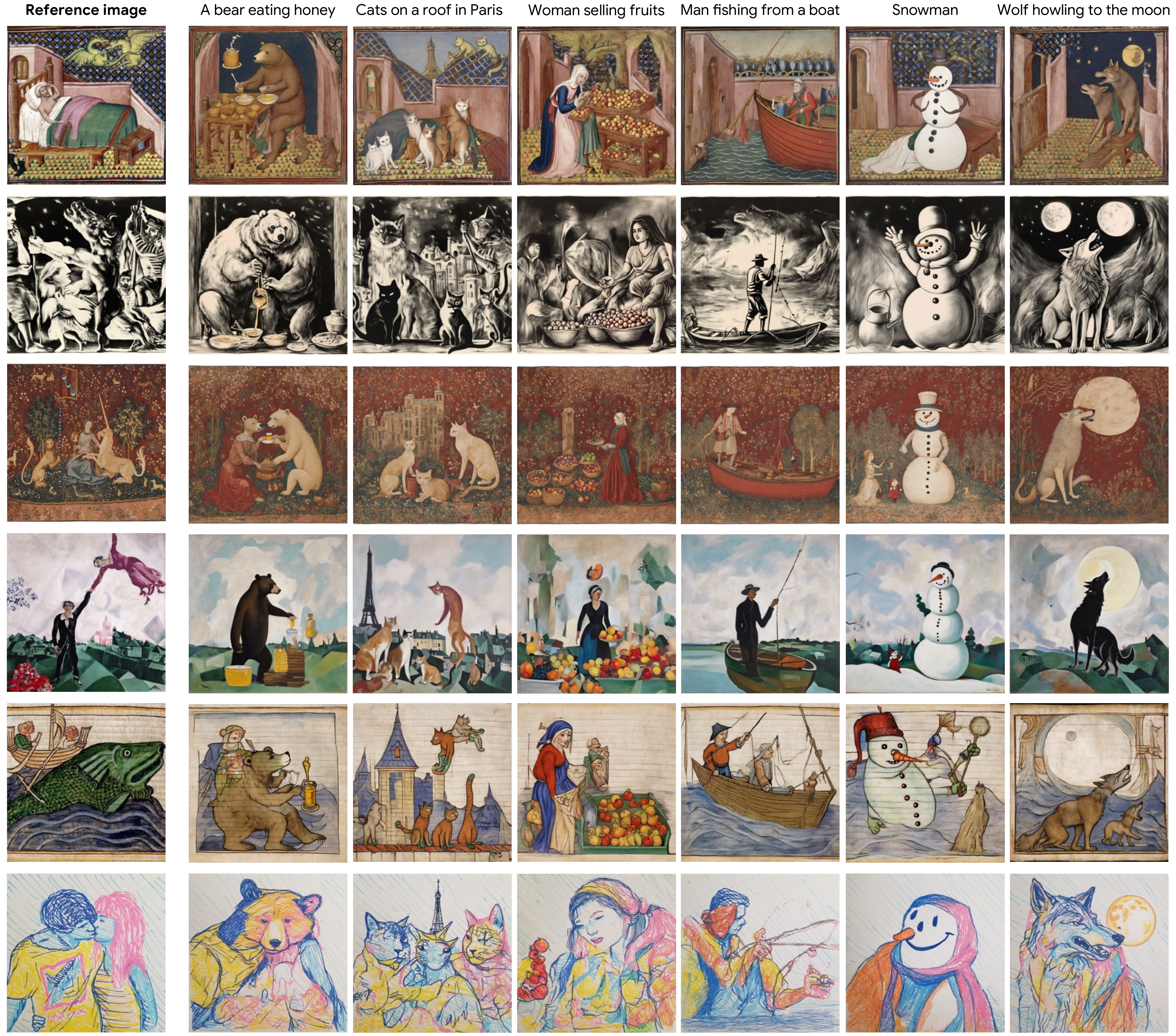}
    \caption{\textbf{Samples of the proposed style transfer techniques applied for a variety of different images and target prompts.}}
    \label{fig:real_grid_2}
\end{figure*}

\begin{figure*}
\vspace{-.25cm}
\footnotesize
\centering
    \includegraphics[width=.98\textwidth]{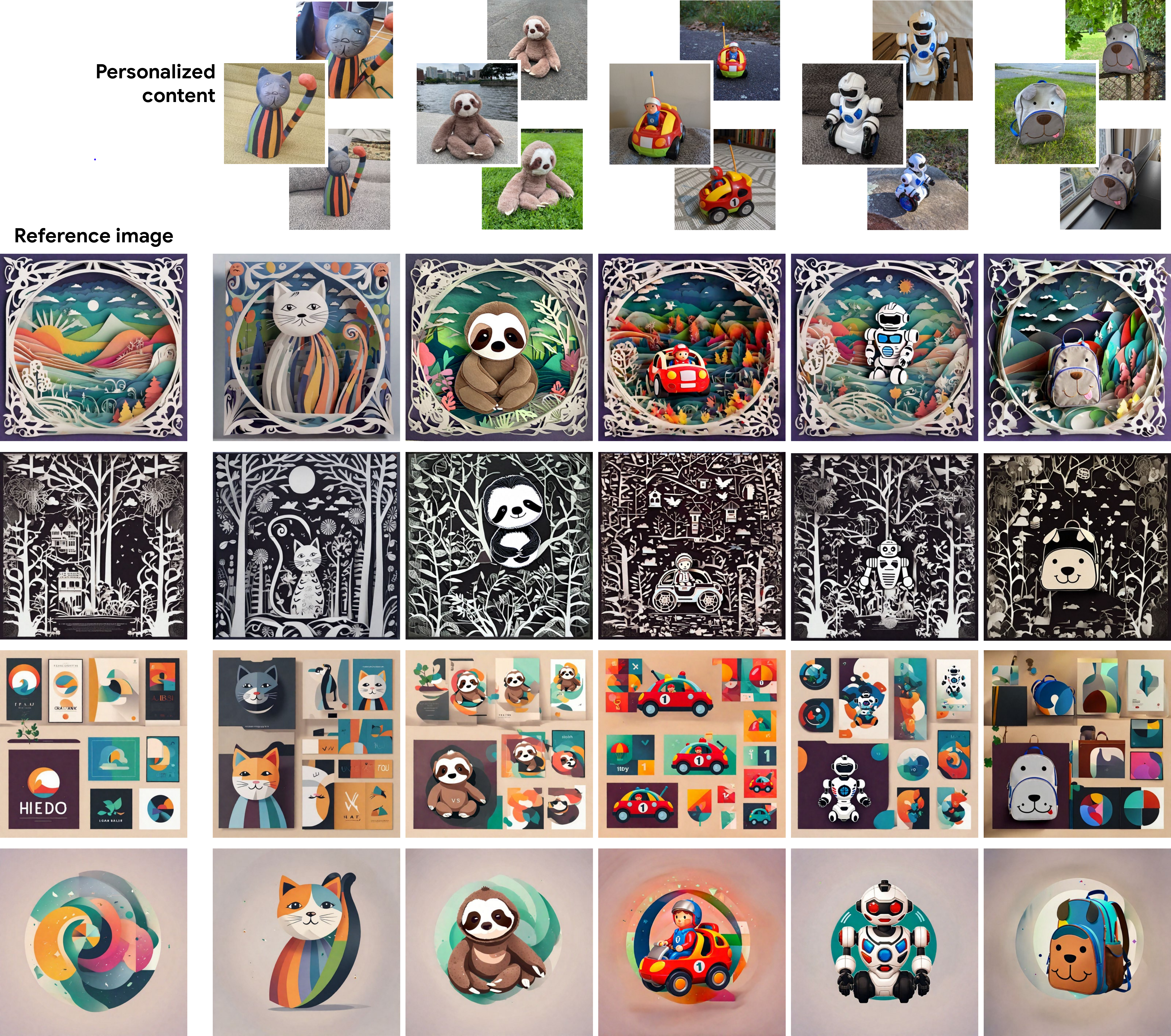}
\caption{\textbf{Personalized T2I diffusion with StyleAligned.} \textit{Each row shows style aligned image st using the reference image on the left, applied on different personalized diffusion models, fine-tuned over the personalized content on top.
The top two rows were generated using the prompt ''[my subject] in the style of a beautiful papercut art.`` The bottom two rows  were generated using the prompt ''[my subject] in beautiful flat design.`` where [my subject] is replaced with the subject name.}} 
\vspace{-0.3cm}
\label{fig:app-db}
\end{figure*}

\begin{figure*}
\vspace{-.25cm}
\footnotesize
\centering
    \includegraphics[width=.91\textwidth]{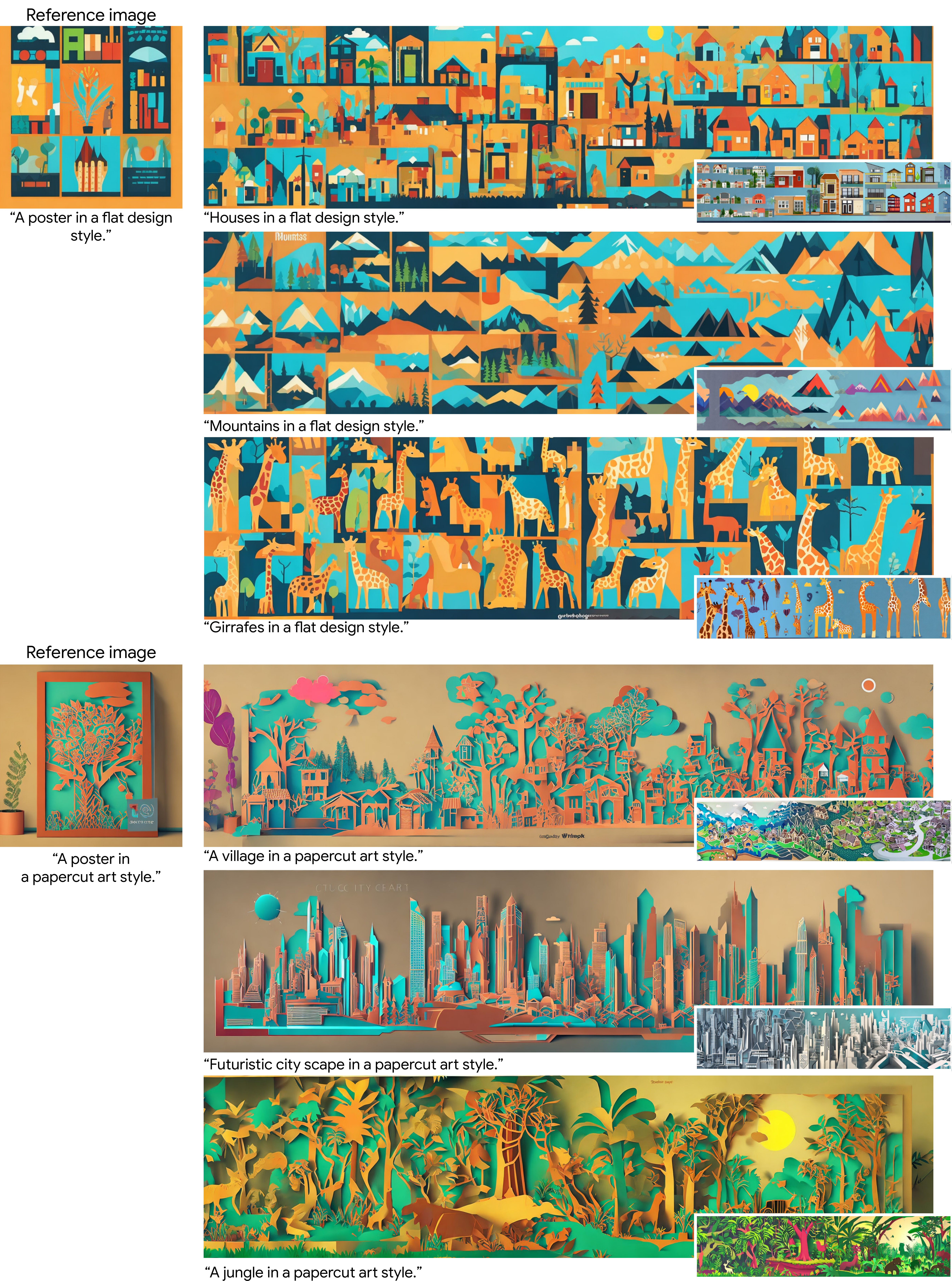}
\caption{\textbf{MultiDiffusion with StyleAligned.} \textit{The panoramas were generated with MultiDiffusion\cite{BarTal2023MultiDiffusionFD} using the text prompt beneath and the left image as reference. The small images in the bottom right corners are the results of MultiDiffusion results without our method. }}
\label{fig:app-multi}
\end{figure*}

\begin{figure*}
\footnotesize
\centering
    \includegraphics[width=1.\textwidth]{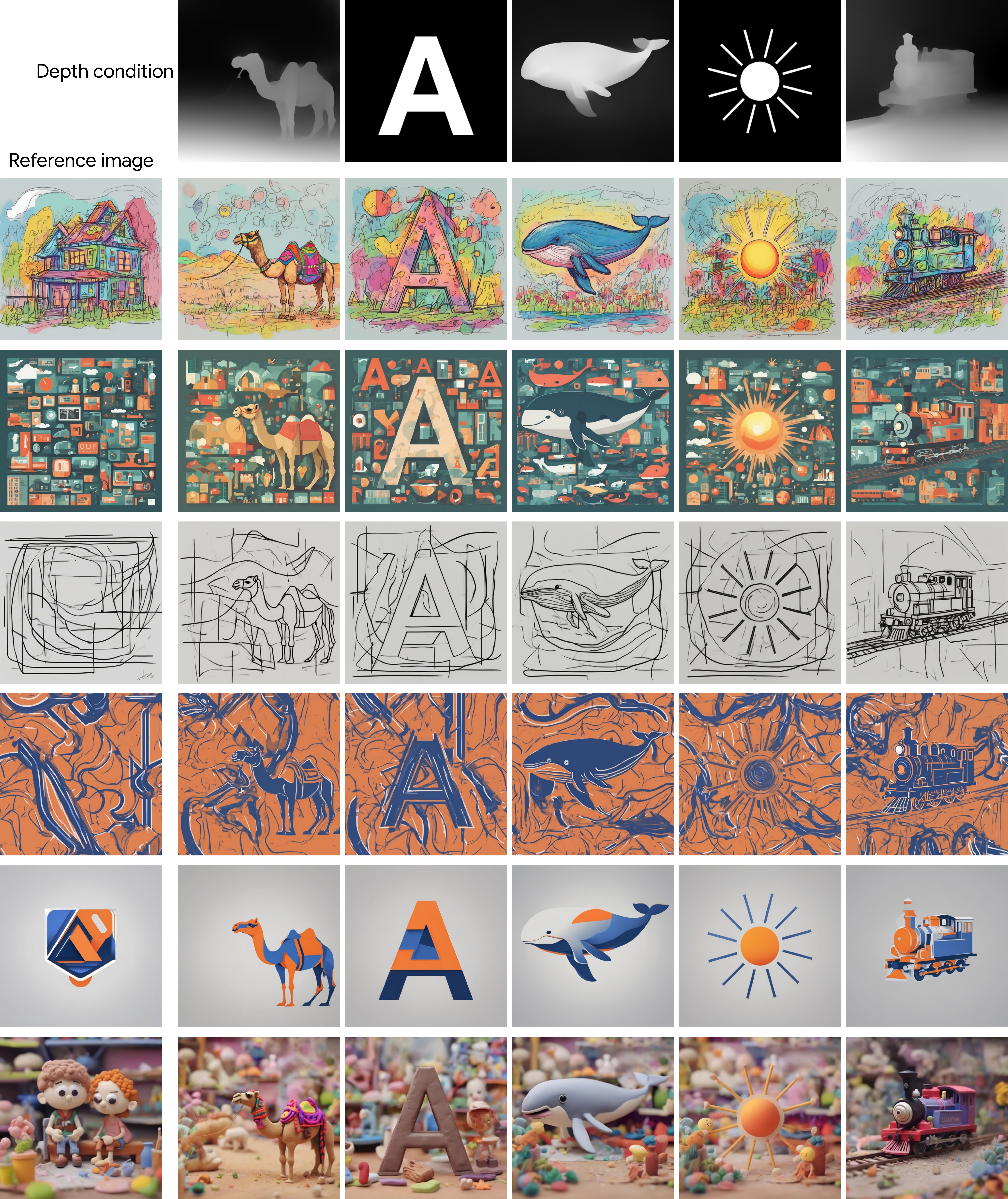}
\caption{\textbf{ControlNet Depth with StyleAligned.}}
\label{fig:app-depth}
\end{figure*}

\begin{figure*}
\footnotesize
\centering
    \includegraphics[width=1.\textwidth]{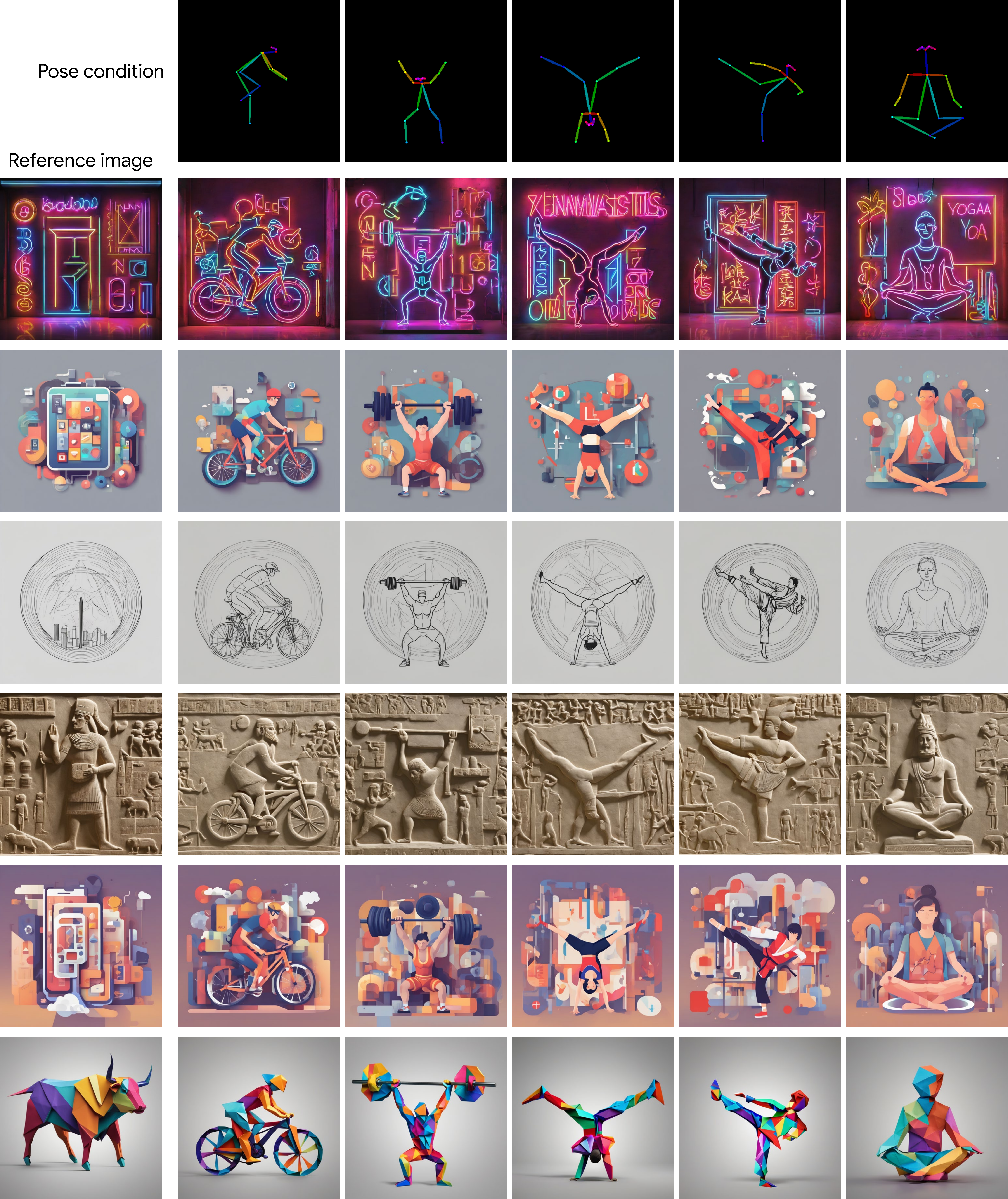}
\caption{\textbf{ControlNet pose with StyleAligned.}}
\label{fig:app-pose}
\end{figure*}

\begin{figure*}[h!]
\footnotesize
\centering
    \includegraphics[width=.97\textwidth]{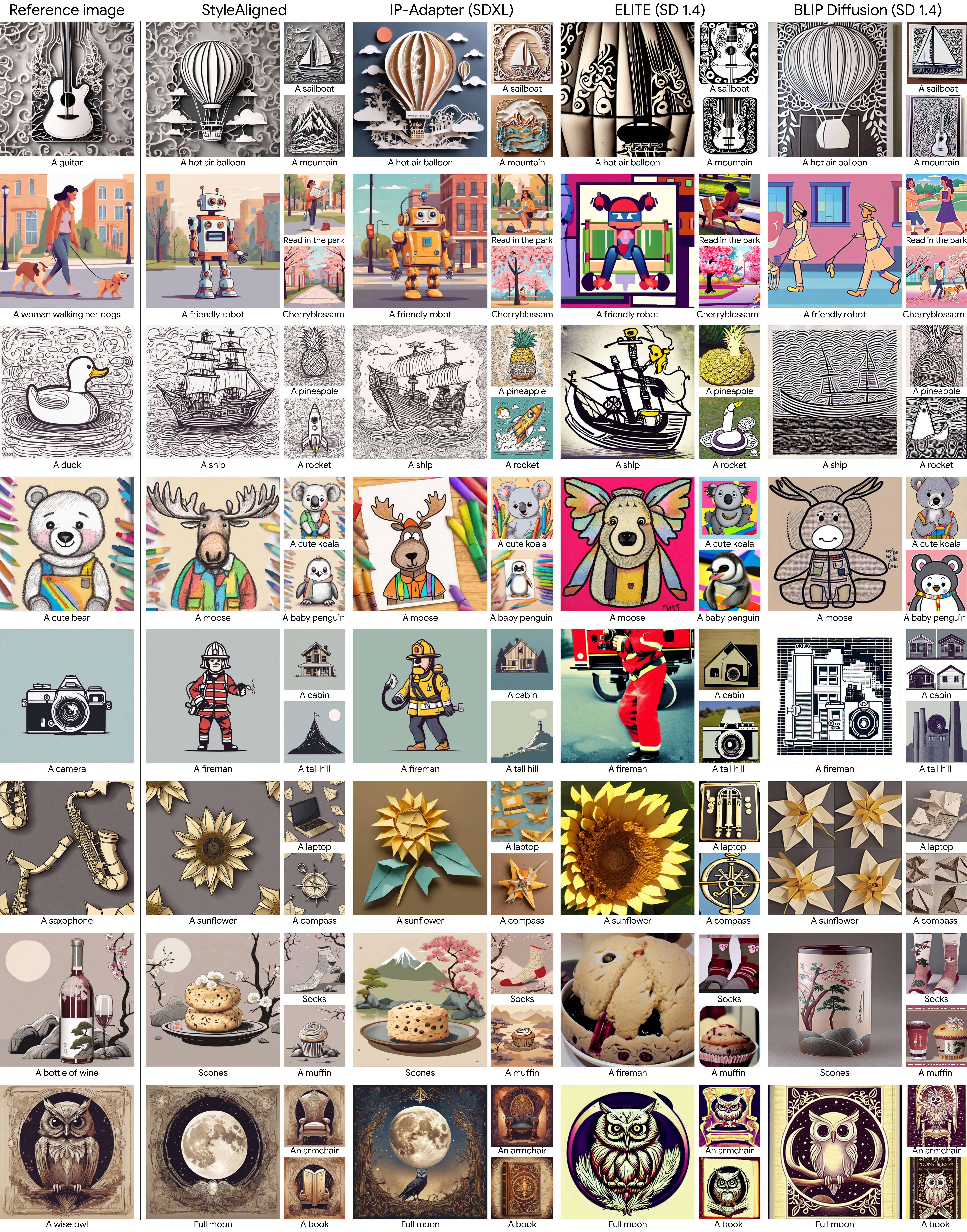}
\caption{\textbf{Qualitative comparison to encoders based personalization methods.}}
\label{fig:comparison-encoders}
\end{figure*}

\begin{figure*}[h!]
\footnotesize
\centering
    \includegraphics[width=.97\textwidth]{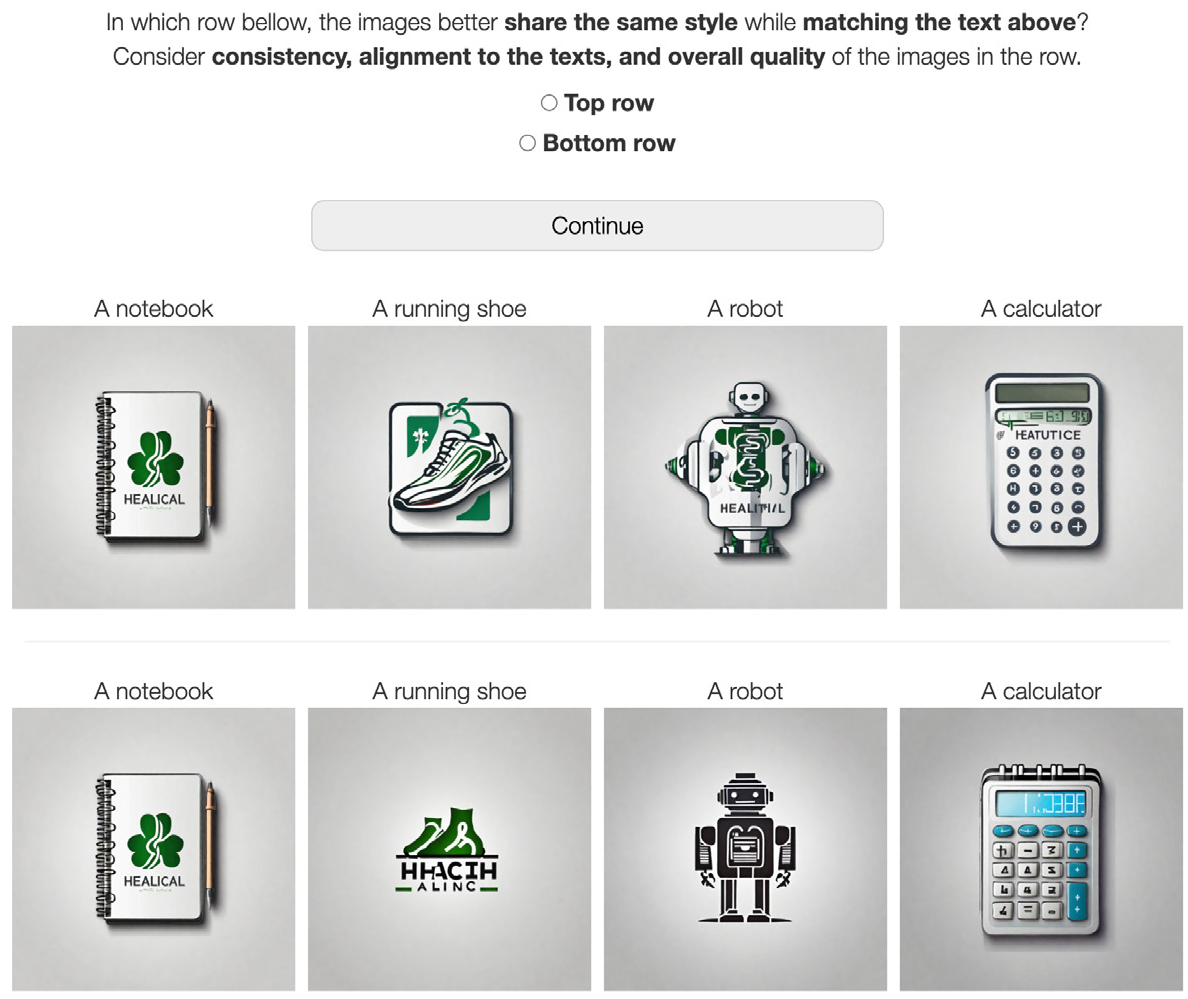}
\caption{\textbf{Screenshot from the user study.} \textit{Each row of images represents the result obtained by different method. The user had to assess which row is better in terms of style alignment and text alignment.}}
\label{fig:screenshot}
\end{figure*}

\clearpage
\onecolumn{
\label{fig:list-prompts}
List of prompts for our evaluation set generation:
\scriptsize{
\begin{lstlisting}[linewidth=\columnwidth,breaklines=true]
1. {A house, A temple, A dog, A lion} in sticker style.
2. {Flowers, Golden Gate bridge, A chair, Trees, An airplane} in watercolor painting style.
3. {A village, A building, A child running in the park, A racing car} in line drawing style.
4. {A phone, A knight on a horse, A train passing a village, A tomato in a bowl} in cartoon line drawing style.
5. {Slices of watermelon and clouds in the background, A fox, A bowl with cornflakes, A model of a truck} in 3d rendering style.
6. {A mushroom, An Elf, A dragon, A dwarf} in glowing style.
7. {A thumbs up, A crown, An avocado, A big smiley face} in glowing 3d rendering style.
8. {A bear, A moose, A cute koala, A baby penguin} in kid crayon drawing style.
9. {An orchid, A Viking face with beard, A bird, An elephant} in wooden sculpture.
10. {A portrait of a person wearing a hat, A portrait of a woman with a long hair, A person dancing, A person fishing} in oil painting style.
11. {A woman walking a dog, A friendly robot, A woman reading in the park, Cherryblossom} in flat cartoon illustration style.
12. {A bithday cake, The letter A,  An espresso machine, A Car} in abstract rainbow colored flowing smoke wave design.
13. {A flower, A piano, A butterfly, A guiter} in melting golden 3d rendering style.
14. {A train, A car, A bicycle, An airplane} in minimalist round BW logo.
15. {A rocket, An astronaut, A man riding a snowboard, A pair of rings} in neon graffiti style.
16. {A teapot, A teacup, A stack of books, A cozy armchair} in vintage poster style.
17. {A mountain range, A bear, A campfire, A pine forest} in woodblock print style.
18. {A surfboard, A beach shack, A wave, A seagull} in retro surf art style.
19. {A paintbrush, A sunflower field, A scarecrow, A rustic barn} in a minimal origami style.
20. {A cityscape, Hovering vehicles, Dragons, Boats} in cyberpunk art style.
21. {A treasure box, A pirate ship, A parrot, A skull} in tattoo art style.
22. {Music stand, A vintage microphone, A turtle, A saxophone} in art deco style.
23. {A tropical island, A mushroom, A palm tree, A cocktail} in vintage travel poster style.
24. {A carousel, Cotton candy, A ferris wheel, Balloons} in retro amusement park style.
25. {A serene river, A rowboat, A bridge, A willow tree} in 3D render, animation studio style.
26. {A retro guitar, A jukebox, A chess piece, A milkshake} in 1950s diner art style.
27. {A snowy cabin, A sleigh, A snowman, A winter forest} in Scandinavian folk art style.
28. {A bowl with apples, A pencil, A big armor, A magical sunglasses} in fantasy poison book style.
29. {A kiwi fruit, A set of d0rums, A hammer, A tree} in Hawaiian sunset painting style.
30. {A guitar, A hot air balloon, A sailboat, A mountain} in papercut art style.
31. {A coffee cup, A typewriter, A pair of glasses, A vintage camera} in retro hipster style.
32. {A board of backgammon, A shirt and pants, Shoes, A cocktail} in vintage postcard style.
33. {A roaring lion, A soaring eagle, A dolphin, A galloping horse} in tribal tattoo style.
34. {A pizza, Candles and roses, A bottle, A  chef} in Japanese ukiyo-e style.
35. {A wise owl, A full moon, A magical chair, A book of spells} in fantasy book cover style.
36. {A cozy cabin, Snow-covered trees, A warming fireplace, A steaming cup of cocoa} in hygge style.
37. {A bottle of wine, A scone, A muffin, Pair of socks} in Zen garden style.
38. {A diver, Bowl of frutis, An astronaut, A carousel} in celestial artwork style.
39. {A horse,A castle,  A cow, An old phone} in medieval fantasy illustration style.
40. {A mysterious forest, Bioluminescent plants, A graveyard, A train station} in enchanted 3D rendering style.
41. {A globe, An airplane, A suitcase, A compass} in travel agency logo style.
42. {A persian cat playing with a ball of wool, A man skiing down the hill, A train at the station, A bear eating honey} in cafe logo style.
43. {A book, A quill pen, An inkwell, An umbrella} in educational institution logo style.
44. {A hat, A strawberry, A screw, A giraffe} in mechanical repair shop logo style.
45. {A notebook, A running shoe, A robot, A calculator} in healthcare and medical clinic logo style.
46. {A rubber duck, A pirate ship, A rocket, A pineapple} in doodle art style.
47. {A trumpet, A fishbowl, A palm tree, A bicycle} in abstract geometric style.
48. {A teapot, A kangaroo, A skyscraper, A lighthouse} in mosaic art style.
49. {A ninja, A hot air balloon, A submarine, A watermelon} in paper collage style.
50. {A saxophone, A sunflower, A compass, A laptop} in origami style.
51. {A penguin, A bicycle, A tornado, A pineapple} in abstract graffiti style.
52. {A magician's hat, A UFO, A roller coaster, A beach ball} in street art style.
53. {A cactus, A shopping cart, A child playing with cubes, A camera} in mixed media art style.
54. {A snowman, A surfboard, A helicopter, A cappuccino} in abstract expressionism style.
55. {A robot, A cupcake, A woman playing bascketball, A sunflower} in digital glitch art style.
56. {A treehouse, A disco ball, A sailing boat, A cocktail} in psychedelic art style.
57. {A football helmet, A playmobil, A truck, A watch} in street art graffiti style.
58. {A cabin, A leopard, A squirrel, A rose} in pop art style.
59. {A bus, A drum, A rabbit, A shopping mall} in minimalist surrealism style.
60. {A frisbee, A monkey, A snake, skates} in abstract cubism style.
61. {A piano, A villa, A snowboard, A rubber duck} in abstract impressionism style.
62. {A laptop, A man playing soccer, A woman playing tennis, A rolling chair} in post-modern art style.
63. {A cute puppet, A glass of beer, A violin, A child playing with a kite} in neo-futurism style.
64. {A dog, A brick house, A lollipop, A woman playing on a guiter} in abstract constructivism style.
65. {A kite surfing, A pizza, A child doing homework, A person doing yoga} in fluid art style.
66. {Ice cream, A vintage typewriter, A pair of reading glasses, A handwritten letter} in macro photography style.
67. {A gourmet burger, A sushi, A milkshake, A pizza} in professional food photography style for a menu.
68. {A crystal vase, A pocket watch, A compass, A leather-bound journal} in vintage still life photography style.
69. {A sake set, A stack of books,  A cozy blanket, A cup of hot cocoa} in miniature model style.
70. {A retro bicycle, A sunhat, A picnic basket, A kite} in outdoor lifestyle photography style.
71. {A group of hikers on a mountain trail, A winter evening by the fire, A hen, A person enjoying music} in realistic 3D render.
72. {A tent, A person knitting, A rural farm scene, A basket of fresh eggs} in retro music and vinyl photography style.
73. {A giraffe, A blanket, A fork and knife, A pile of candies} in cozy winter lifestyle photography style.
74. {A wildflower, A ladybug, An igloo in antarctica, A person running} in bokeh photography style.
75. {A coffee machine, A laptop, A person working, A plant on the desk} in minimal flat design style.
76. {A camera, A fireman, A wooden house, A tall hill} in minimal vector art style.
77. {A person texting, A person scrawling, A cozy chair, A lamp} in minimal pastel colors style.
78. {A smartphone, A book, A dinner table, A glass of wine} in minimal digital art style.
79. {A brush, An artist painting, A girl holding umbrella, a pool table} in minimal abstract illustration style.
80. {A pair of running shoes, A motorcycle, Keys, A fitness machine} in minimal monochromatic style.
81. {A compass rose, A cactus, A zebra, A blizzard} in woodcut print style.
82. {A lantern, A tricycle, A seashell, A swan} in chalk art style.
83. {Magnifying glass, Gorilla, Airplane, Swing} in pixel art style.
84. {Hiking boots, Kangaroo, Ice cream cone, Hammock}  in comic book style.
85. {Horseshoe, Vintage typewriter, Snail, Tornado} in vector illustration style.
86. {A lighthouse, A hot air balloon, A cat, A cityscape} in isometric illustration style.
87. {A compass, A violin, A palm tree, A koala} in wireframe 3D style.
88. {Beach umbrella, Rocket ship, Fox, Waterfall} in paper cutout style.
89. {Tree stump, Harp, Chameleon, Canyon} in blueprint style.
90. {Elephant, UFO toy, Flamingo, Lightning bolt} in retro comic book style.
91. {Robot, Temple, Jellyfish, Sofa} in infographic style.
92. {Microscope, Giraffe, Laptop, Rainbow} in geometric shapes style.
93. {Teapot, Dragon toy, Skateboard, Storm cloud} in cartoon line drawing style.
94. {Crystal ball, Carousel horse, Hummingbird, Glacier} in watercolor and ink wash style.
95. {Feather quill, Satellite dish, Deer, Desert scene} in dreamy surreal style.
96. {Map, Saxophone, Mushroom, Dolphin} in steampunk mechanical style.
97. {Anchor, Clock, Globe, Bicycle} in 3D realism style.
98. {Clock, Helicopter, Whale, Starfish} in retro poster style.
99. {Binoculars, Bus, Pillow, Cloud} in bohemian hand-drawn style.
100. {Rhino, Telescope, Stool, Panda} in vintage stamp style.
\end{lstlisting}
}
}


\end{document}